\pdfoutput=1

\documentclass[11pt]{article}

\usepackage[preprint]{acl}

\usepackage{amsmath}
\usepackage{amsfonts}
\usepackage{amssymb}
\usepackage{graphicx}
\usepackage{booktabs}
\usepackage{caption}
\usepackage{changepage}
\usepackage{times}
\usepackage{latexsym}
\usepackage{multicol}
\usepackage{tabularx}
\usepackage[T1]{fontenc}
\usepackage[T1]{fontenc}
\usepackage{inconsolata}  
\usepackage{upquote}
\usepackage{newtxtext}
\usepackage{courier}
\usepackage{makecell}

\usepackage{silence}
\usepackage{subcaption}
\usepackage[ruled,vlined,linesnumbered]{algorithm2e} 
\usepackage{amsmath,amssymb}

\usepackage{comment}
\usepackage[T1]{fontenc}

\usepackage[utf8]{inputenc}

\usepackage{microtype}

\usepackage{inconsolata}
\usepackage{booktabs}     
\usepackage{siunitx}      
\usepackage{graphicx}
\usepackage[most]{tcolorbox}  

\usepackage{enumitem}
\usepackage{fancyhdr}
\usepackage{url} 

\title{Can Lessons From Human Teams Be Applied to Multi-Agent Systems? \\The Role of Structure, Diversity, and Interaction Dynamics}

\author{
  Rasika Muralidharan\thanks{\,Corresponding author: \texttt{rasimura@iu.edu}}, 
  Haewoon Kwak,
  Jisun An \\
  Indiana University Bloomington \\
  \texttt{\{rasimura, hwkwak, jisunan\}@iu.edu}
}

\tcbuselibrary{breakable}
\begin{document}
\maketitle

\begingroup
\renewcommand\thefootnote{\textasteriskcentered}
\footnotetext{\,Our repository, including code and data, is available at \url{https://github.com/Rasikamurali/ai-agents}.}
\endgroup

\begin{abstract}

Multi-Agent Systems (MAS) with Large Language Model (LLM)-powered agents are gaining attention, yet fewer studies explore their team dynamics. Inspired by human team science, we propose a multi-agent framework to examine core aspects of team science: structure, diversity, and interaction dynamics. We evaluate team performance across four tasks: CommonsenseQA, StrategyQA, Social IQa, and Latent Implicit Hate, spanning commonsense and social reasoning. Our results show that flat teams tend to perform better than hierarchical ones, while diversity has a nuanced impact. 
Interviews suggest agents are overconfident about their team performance, yet post-task reflections reveal both appreciation for collaboration and challenges in integration, including limited conversational coordination.
\end{abstract}

\section{Introduction}

Large Language Models (LLMs)' growing ability to process, generate, and reason with natural language has driven interest in designing multi-agent systems (MAS)---collections of AI agents collaborating on complex problems. These systems offer several advantages: supporting distributed problem-solving, representing diverse viewpoints, and simulating collaborative dynamics such as debate, negotiation, and cooperation~\cite{du_improving_2023, chen_reconcile_2024, li_more_2024, zhu_multiagentbench_2025, zhang_cumulative_2024, wang_talk_2025}. MAS allows us to explore social phenomena and study interaction dynamics that mirror human team behavior. However, few studies examine agent structures, diversity effects, and interactions, despite their potential efficiency and adaptability~\cite{wu2025hiddenstrengthdisagreementunraveling, bettini2025impactbehavioraldiversitymultiagent, celebrate_diversity}. Well-designed structure and diversity can also foster trust and alignment in human-AI collaboration~\cite{stahl2021unraveling, delice2019advancing, hattori2023closing, mcgrath2024collaborativehumanaitrustchait}.

\looseness=-1
In addition, recent work on human-AI collaboration highlights the critical role of coordination and communication~\cite{stahl2021unraveling, yang2024llmbasedmultiagentsystemstechniques, agashe-etal-2025-llm, li_metaagents_2023}.
Building on this, we draw on insights from human team science to better understand collaboration in AI teams.
This literature emphasizes the importance of team structure, including how authority and communication are organized, and diversity in members' backgrounds and perspectives.
It further stresses that effective collaboration depends not only on outcomes but also on how team members \textit{understand}, \textit{coordinate}, and \textit{reason} together.

This leads to our central question: \textit{Can principles from team science help us design more effective AI teams?} 
To explore this, we ground our study in theories from organizational science. Prior work suggests that flat team structures encourage open communication and trust, while hierarchical structures can expedite decision-making through defined roles~\cite{ji_how_2020, greer_why_2018}. Diversity-performance theory further suggests that teams with diverse backgrounds can outperform homogeneous ones by bringing in broader perspectives~\cite{cox_managing_1991, pelled_exploring_1999, van_knippenberg_synergy_2020}.
Building on these foundations, we propose three research questions:
\begin{itemize}[labelindent=0pt, leftmargin=*, noitemsep, topsep=0pt]
    \item \textbf{RQ1 (Structure)}:  How does team structure (flat versus hierarchical) affect team performance across reasoning and inference tasks?
    \item \textbf{RQ2 (Diversity):} How does demographic diversity, instantiated via agent personas, influence team performance, and does its impact vary by team structure?
    \item \textbf{RQ3 (Interaction):} How do agents perceive their roles and interactions within the team, and what do their communication patterns reveal about coordination, understanding, and reasoning?
\end{itemize}

To address these questions, we simulate flat and hierarchical teams of LLM agents, each assigned demographic personas (e.g., age, race, gender, occupation), and evaluate them on four tasks requiring reasoning, social inference, and normative judgment: CommonsenseQA~\cite{commonsenseQA}, StrategyQA~\cite{strategyQA}, Social IQa~\cite{social_iqa}, and Latent Implicit Hate Detection~\cite{implicit_hate}. These tasks are chosen for their reliance on nuanced reasoning and value-sensitive judgment, which are shaped by team structure and diversity. Beyond benchmark evaluations, for assessing the quality and nuance of the interactions, we employ an `LLM-as-a-judge'~\cite{gu2024survey} approach.  We calibrate our LLM judge with a small, human-annotated dataset using few-shot prompting for efficient and reliable assessment.  

\begin{figure*}[t]
    \centering
    \includegraphics[width=140mm, height=65mm]{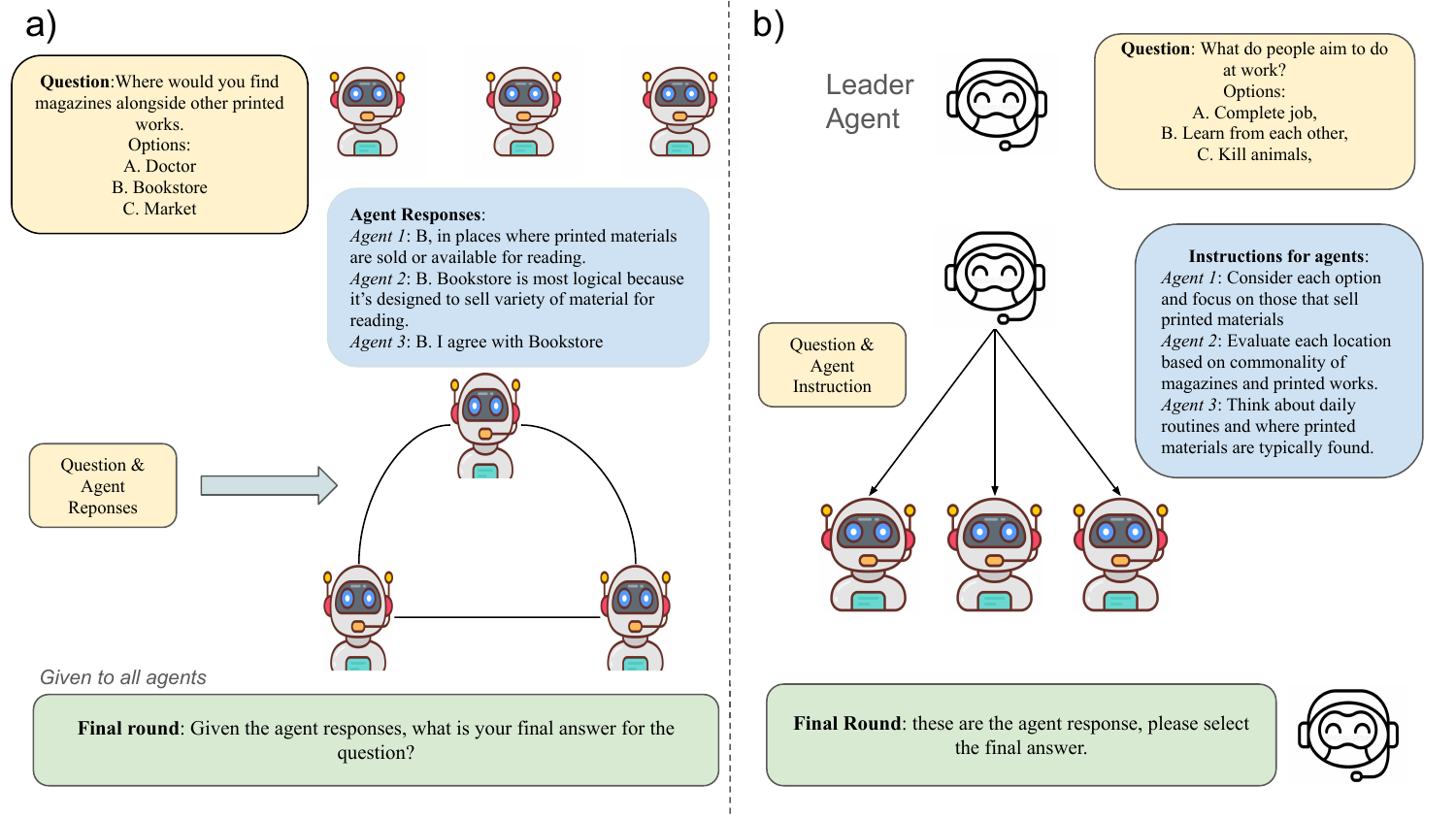}
    \caption{Conversation flows in (a) flat  and (b) hierarchical teams. In flat teams, agents respond independently and iteratively refine their answers. In hierarchical teams, leader agents issue instructions and determine the final answer based on others' responses.}
    \label{fig:conversation_flow}
\end{figure*}

Together, this study offers a theory-driven investigation of how structure and diversity shape both performance and internal dynamics of AI teams. 
Our findings show that these dimensions significantly impact how agents interact, reason, and coordinate. This, in turn, offers design insights for building more interpretable, collaborative, and socially aware AI teams.
Our contributions are:     
\begin{itemize}[labelindent=0pt, leftmargin=*, noitemsep, topsep=0pt]
    \item A framework for multi-agent LLM teams to systematically evaluate core principles from team science, such as team structure and team diversity.
    \item A comprehensive evaluation of team performance and interaction through quantitative performance and qualitative interaction analysis.
    \item Empirical findings on how team structure and composition affect reasoning and social inference.
    \item Theoretical implications for MAS with LLM design: the mediating roles of communication structure and social framing in reasoning and coordination.
\end{itemize}

\section{Background}

\subsection{Multi-Agent Frameworks for LLMs}

MAS are collections of intelligent agents that interact in a shared environment to achieve individual and collective goals. A defining feature is interaction---the ability to communicate, coordinate, and negotiate to accomplish tasks. In Natural Language Processing (NLP), MAS enable advanced reasoning and social understanding~\cite{hegazy_diversity_2024, wang_contextualized_2023, xu2023reasoninglargelanguagemodels}.  

A prominent paradigm is multi-agent debate, where multiple LLMs engage in structured argumentation to improve factual accuracy, identify reasoning failures, and simulate consensus~\cite{chen_reconcile_2024, du_improving_2023, liang_encouraging_2024}. Another line of work explores hierarchical teams, modeling organizational structures with chains of command and task delegation~\cite{wang_talk_2025, zhu_multiagentbench_2025}. A complementary trend assigns social characteristics to agents, such as personality traits or demographics, to study emergent behaviors. Studies have shown that 
MAS with LLMs exhibit human-like social phenomena when social characteristics are incorporated
~\cite{park_generative_2023, chuang_simulating_2024, zhang_exploring_2024, chuang_beyond_2024, chen_persona_2024, jiang_personallm_2024, samuel_personagym_2024, park_predict}.

\subsection{Insights from Team Science}

To guide our investigation, we turn to team science, a multidisciplinary field that examines the factors driving effective collaboration. Decades of research emphasize two key determinants of team performance: structure and diversity~\cite{ji_how_2020, evans_flatteam_innovation, cooke_enhancing_2015, effects_teamdiversity, salas2008teamwork, cox_managing_1991}. 

\noindent \textbf{Team structure} affects information flow, decision-making, and conflict resolution~\cite{hackman2002leading,salas2008teamwork}. Two common structures are: (1) flat, with decentralized decision-making, which fosters openness but can lack accountability and scalability and (2) hierarchical, with clear authority layers, which improves coordination but risks communication silos~\cite{greer_why_2018}.

\noindent \textbf{Team diversity}, encompassed in demographic, cognitive, and functional differences, can enhance team efficacy. Diverse teams are often more innovative~\cite{effects_teamdiversity, van_knippenberg_synergy_2020}, avoid cognitive traps, and excel in logical reasoning and social inference~\cite{roberge_recognizing_2010}. Yet, they may also face communication barriers and increased conflict~\cite{cox_managing_1991}.

\noindent \textbf{Interaction dynamics}, such as communication patterns, coordination mechanisms, and leadership styles, are essential for team success. Research shows that trust calibration, role negotiation, and adaptive communication significantly shape outcomes in both human and human-AI teams~\cite{stahl2021unraveling}. 

LLM-based MAS provide a promising testbed to explore how structure, diversity, and interaction dynamics affect performance on NLP tasks. While MAS have been used in commonsense and social reasoning, few studies systemically compare the effects of structure and diversity on team outcomes.

\section{Multi-Agent Team Design}

Team science identifies structure and diversity as key to human collaboration. We operationalize these theoretical constructs into multi-agent design, examining how structure, diversity, and communication shape AI team behavior and effectiveness.

\subsection{Team Structure (RQ1)}\label{team structure}

Flat and hierarchical structures are two central organization forms in team science. 
In our study, flat teams consist of 3, 5, or 7 agents, odd numbers to enable majority voting without ties. Teams engage in a 2-4 round debate. As shown in Fig.~\ref{fig:conversation_flow} a), 
agents first provide independent responses (Round 0), stored in shared memory.
In subsequent rounds, they review and revise their answers while considering others' input. In the final round, each agent submits a final judgment, and the team decision is determined by majority vote.

Hierarchical teams follow a top-down communication structure, with designated leaders who delegate tasks and synthesize responses. We implement two variants: 1) a 4-agent team with one leader and three subordinates; 2) a 7-agent team with one leader, two managers, and four subordinates (two per manager). 

As shown in Fig~\ref{fig:conversation_flow} b), in Round 0, the leader receives the question and issues tailored instructions to each agent, simulating division of labor, specialization, and perspective diversification.
These instructions guide how agents interpret the question, which aspect to prioritize, and what reasoning strategy to apply (e.g., focus on edge cases, consider the most probable answer first, and identify counterexamples). 
In the 7-agent setting, the leader provides meta-instructions to managers (e.g., gather diverse reasoning paths or probe conflicting assumptions), who relay directives to subordinates. 
Agents respond accordingly, and their outputs are returned to the leader. In later rounds, the leader refines guidance or resolves inconsistencies, and in the final round, reviews all inputs to makes the team's decision, potentially overriding the majority to exercise hierarchical veto power. 

\subsection{Team Diversity (RQ2)} \label{team diversity}

Team science  emphasizes the role of diversity,  particularly in demographic and experiential attributes, as a key determinant of team performance. To examine its effect on reasoning and coordination in LLM-based teams, we assign each agent a persona that reflects  human demographics and systematically test teams with varying compositions.

Each persona is defined along four dimensions: gender (male, female), age (young adult, young working professional, working professional, senior), ethnicity (White, Black, Asian), and occupation (white- or blue-collar). These dimensions are well-established markers of social identity known to influence communication, authority, and decision-making in human teams~\cite{kunze2022age, joshi2009role, song2020comparing}.




\subsection{Interaction Dynamics (RQ3)} \label{interaction dynamics}

Team science often uses pre- and post-interviews of participants as a diagnostic tool to understand the how and the why of shifts of observed changes in experiment settings~\cite{Martina2010, Hamilton2019Qualitative}. Particularly, the pre-interview establishes a baseline of expectation or pre-existing team-norms, while post-interview evaluates the perceived effectiveness of a change and the shift in dynamics. 
The pre- and post-interviews have been use for process tracing, and providing evidence needed to reveal causal changes by which team structure and diversity influence interactions over time~\cite{stahl2021unraveling, Wood2024Using}. 
As LLMs become more deeply embedded in our society, there is a greater likelihood that LLMs will begin interacting with each other~\cite{park_generative_2023, Moller2025Impact, Coppolillo2024Engagement}.
Therefore, it is essential to assess the priors of these representative models. To that extent, we adopt and modify the "interviews" from team science as an "elicitation probing" method~\cite{Bo2024Reflective, Wang2024Devils, Kim2025ReflAct}.

To capture these aspects in multi-agent settings, we first incorporate pre- and post-task elicitation probing. Specifically, we assess understanding of team goals, perceived role clarity, and reasoning process. The qualitative feedback complements our quantitative measures and offers deeper insights into intra-team coordination. This method serves as a prompt-based inquiry to show the differences in context-conditioned outputs. Differences between pre- and post-responses can be interpreted as agents' expressed task priors. 

\noindent We ask five questions to gauge initial priors about team goals, individual role, individual readiness, and perceived inclusiveness. The full set of questions are available in Appendix~\ref{pre-task}. After the task, we ask six question as part of the post-elicitation querying to assess how the team experience may have shifted perceptions~\ref{post-task}. Together, these elicitation probings offer a window into shifts in confidence, role clarity, and perceived synergy. This reflective process helps assess how well agents align in comprehension and coordination. 

While the elicitation responses offer some insight into agent interaction, they do not fully capture the quality of agent-to-agent communication. To better evaluate these dynamics, we adopt an \emph{LLM-as-a-judge} approach, using GPT-4o to score sample team conversations across five dimensions. To ensure the LLM's assessments align with human judgment, we calibrated its scoring criteria on a subset of human-rated conversations (\S5.3.1). Each conversation is rated on a 1-5 scale (5 = highest):
\begin{enumerate}[label=Q$^{\text{judge}}_\arabic*.$, labelindent=0pt, leftmargin=*, noitemsep, topsep=0pt]
    \item How well do the agents understand each other and collectively complete the task?
    \item How well do the agents coordinate, delegate tasks and integrate ideas?
    \item How strong is the team's reasoning compared to what an individual agent might produce?
    \item How clear, coherent and logically structure is the conversation? 
    \item How confident are you in the team's final answer based on their reasoning?
\end{enumerate}

\noindent This provides a complementary view of how agents engage, reason together, and coordinate toward shared goals, beyond what is captured in elicitation responses or performance metrics. 

\section{Experiment settings}



Implementation details, including prompt designs, are provided in the Appendix~\S \ref{appendix section: implementation}. To ensure reproducibility, we use four open-source LLMs: Meta’s LLaMA-8B Instruct, Alibaba’s Qwen-7.5B Instruct, Mistral-7B v0.3 Instruct, and DeepSeek R1-8B. Given, we prioritize effects of team science elements, rather than maximizing team performance, 7B-8B models provide sufficient linguistic competence. Additionally, 7B-8B models are present across all the model families, making the comparison more robust.

\subsection{Datasets}
Our evaluation leverages four datasets. CommonsenseQA~\cite{commonsenseQA} (CS), a multiple-choice dataset testing general common sense, and StrategyQA~\cite{strategyQA} (ST), which requires strategic reasoning over a knowledge graph, assess agents' commonsense understanding. In contrast, Social-IQa~\cite{social_iqa} (SQA), which focuses on reasoning about social interactions and motivations, and Implicit Hate dataset~\cite{implicit_hate} (IH), designed to identify subtle forms of hate speech, evaluate agents' social reasoning in nuanced contexts. We focus on commonsense and social reasoning tasks as these domains are foundational to effective team coordination, requiring agents to interpret context, anticipate others’ perspectives, and reason beyond surface-level cues. In human teams, the ability to perform these actions relies on cognitive and interpersonal skills that are known to be shaped by team structure and dynamics~\cite{cox_managing_1991}.

For brevity, we refer to these datasets using their abbreviations (CS, ST, SQA, IH) in all subsequent tables and figures.

\subsection{Team Structure Experiments}  
 
We evaluate team structure by comparing bootstrapped accuracy between flat and hierarchical teams  across datasets.
For this comparison, we use the full test or validation sets of CommonsenseQA, StrategyQA, and Social IQa. For Implicit Hate, we use the stage 1 set of data, which labels each post as `implicit hate,' `explicit hate,' or `non-hate.' We sample 500 from each class to ensure balance and match the overall scale of the other three datasets. 

\subsection{Team Diversity Experiments}  
To evaluate the impact of demographic diversity on team performance, we compare persona-based teams, where diversity is introduced through assigned personas, with matched no-persona teams, across both flat and hierarchical structures. In persona-based teams, each agent is assigned a persona along four demographic dimensions (e.g., age, gender, ethnicity, occupation), introducing controlled diversity into the team composition.

For each experimental configuration (model, task, rounds), we match team size and structure between conditions. We then conduct paired statistical tests (paired $t$-tests and Wilcoxon signed-rank tests) and compute Cohen's $d$ and mean accuracy deltas to assess significance and effect size. 

To quantify team diversity, we use Gini's Index~\cite{gini_index}, which captures variation across demographic dimensions. 
As exhaustively testing all persona combinations across team sizes is infeasible, we apply stratified sampling to generate teams and select 15 combinations per setting, with an equal number of high, medium, and low diversity teams.
Intuitively, high-diversity teams feature agents with maximal differences across the four persona dimensions, while low-diversity teams consist of agents with mostly overlapping demographic traits.
This allows us to systematically study the impact of team diversity on task performance.
Diversity experiments are run on a 10--20\% random subsample of the test or validation sets. 
Teams without personas are evaluated on the same subsample for consistency. 
To confirm robustness, we additionally test the best-performing model on the full test or validation datasets. 
For the Implicit Hate dataset, we sample 100 examples each from the `implicit hate,' `explicit hate,' and `non-hate' categories to maintain class balance.

\section{Results}
\subsection{Effect of Team Structure on Performance}

Across all models and tasks, flat teams consistently outperform hierarchical ones, as shown in Table~\ref{tab:flat_vs_hier_singlecol}. A paired $t$-test over all comparisons confirms the significance of this difference ($t = 2.6230$, $p = 0.0192$), with an average performance gain of 5.26 points in favor of flat teams.

\begin{table}[t!]
\centering
\caption{Accuracy comparison of teams across models and tasks. Scores are flat/ hierarchical respectively.}
\begin{tabular}{lcccc}
\toprule
\textbf{Model} & \textbf{CS} & \textbf{ST} & \textbf{SQA} & \textbf{IH} \\
\midrule
DeepSeek & 66 / 50 & 61 / 55 & 49 / 42 & 38 / 32 \\
LLaMA    & 79 / 69 & 67 / 51 & 54 / 44 & 44 / 39 \\
Mistral  & 71 / 64 & 63 / 57 & 52 / 45 & 36 / 41 \\
Qwen     & 85 / 75 & 61 / 52 & 68 / 54 & 49 / 42 \\
\bottomrule
\end{tabular}
\label{tab:flat_vs_hier_singlecol}
\end{table}

\begin{table}[t!]
\centering
\caption{Paired t-test comparing flat vs. hierarchical team performance (no diversity condition) across tasks. Significance: * $p<0.05$, ** $p<0.01$, *** $p<0.001$}
\begin{tabularx}{\columnwidth}{Xrrr}
\toprule
\textbf{Task} & \textbf{t-stat} & \textbf{Mean Diff.} & \textbf{Cohen's d} \\
\midrule
CS    & 2.69$^*$   & 9.54  & 1.35 \\
ST    & 5.13$^*$   & 5.89 & 2.18 \\
SQA   & 0.53    & 0.89  & 0.26 \\
IH    & -0.35   & -1.38 & -0.18 \\
\bottomrule
\end{tabularx}
\label{tab:team_structure_bytask}
\end{table}

To assess whether this structural advantage varies by task, we conduct paired $t$-tests on each dataset individually (Table~\ref{tab:team_structure_bytask}). Flat teams significantly outperform hierarchical teams on {StrategyQA} ($t = 4.36^*$, $d = 2.18$) and {CommonsenseQA} ($t = 2.70^*$, $d = 1.35$). In contrast, the differences for {Social IQa} and {Implicit Hate} are small and not statistically significant, suggesting that the effect of team structure may be task-dependent.

These findings indicate that flat teams are especially well-suited for tasks requiring procedural reasoning or multi-step inference, such as strategy problems. In such settings, the peer-to-peer nature of flat communication likely enables more efficient information exchange and decision convergence. Conversely, hierarchical structures may introduce information bottlenecks or distortion as messages propagate across layers, diminishing responsiveness and fidelity, particularly detrimental in tasks where contextual nuance is crucial.


\subsection{Effect of Team Diversity on Performance}

This section analyzes how task accuracy is affected by demographic diversity, comparing persona-based and no-persona teams, and examining performance variation by Gini-based diversity levels across four tasks.

Across all experimental pairs, we observe a statistically significant performance decline in flat teams when diversity is introduced (t-test = $-14.86$, Cohen's $d = -0.21$, $p < 0.05$) , with an average drop of 1.35\% points.  This may stem from increased conflict or misalignment in communication, as agents reason from different demographic perspectives via assigned personas. Hierarchical teams also show a small but significant decline (t = $-2.76$, Cohen's $d = -0.06$, $p < 0.001$), suggesting that structured communication may limit the effective use of demographic cues. On average, hierarchical teams experience a 0.3\% point drop in performance with the addition of personas. 

\begin{table}[ht]
\small
\centering
\caption{Paired $t$-test values comparing diversity vs. no-diversity 
(Flat = Flat (3 agents), r. = rounds, Hier. = Hierarchical. 
All results are significant at $p < 0.05$)}
\begin{tabular}{lcccc}
\toprule
\textbf{Setting} & \textbf{CS} & \textbf{ST} & \textbf{SQA} & \textbf{IH} \\
\midrule
Flat, 2 r.     & -15.43$^*$ & -3.13$^*$ & -15.01$^*$ & 8.58$^*$  \\
Flat, 3 r.     & -12.6$^*$  & -1.16$^*$ & -15.46$^*$ & 2.01$^*$  \\
Flat, 4 r.     & -13.04$^*$ & -2.3$^*$  & -15.17$^*$ & 6.96$^*$  \\
Hier., 2 r.    & -1.46$^*$  & -0.25$^*$ & 3.5$^*$    & -4.3$^*$  \\
Hier., 3 r.    & 1.25$^*$   & -2.9$^*$  & 4.18$^*$   & -3.48$^*$ \\
Hier., 4 r.    & -0.78$^*$  & -2.81$^*$ & 3.98$^*$   & -7.68$^*$ \\
\bottomrule
\end{tabular}
\label{tab:ttest_tasks}
\end{table}

We further investigate the relationship between diversity and team performance across different team settings, as shown in Table~\ref{tab:ttest_tasks}. 
Flat teams consistently show significant performance declines with diversity, with large negative effect sizes (Cohen's $d$ ranging from $-0.56$ to $-0.84$) and t-statistics between 13 and 16. Hierarchal teams maintain the overall negative trend. The results highlight that the impact of demographic personas varies by team structure: flat teams are more sensitive to composition, showing both stronger gains and sharper declines. In contrast, hierarchical teams exhibit weaker and more inconsistent effects. This suggests that open communication structures may magnify the influence of diversity, depending on how well team members align.

We extend this analysis to task level, to assess if the effects of diversity are task-dependent. For each dataset, team performance is evaluated as a function of diversity level, quantified using the Gini index (see Appendix~\ref{appendix section: team div}). Task-specific trends indicate that diversity's impact is shaped by both team composition and task characteristics, rather than reflecting a task-independent effect. For example, CommonsenseQA exhibits a modest but steady increase in average accuracy for flat teams as diversity rises, suggesting a consistent benefit from diverse perspectives. In contrast, Implicit Hate task demonstrates increased variance, especially in hierarchical teams, where some configurations excel while others fail to coordinate. While diversity often hinders performance, tasks requiring social reasoning and normative understanding may benefit from aligned persona perspectives. This variance-amplifying effect echoes findings from human team science~\cite{van2004work}, which suggest that diversity tends to increase the spread of outcomes rather than ensuring improvement. These sensitivities highlight the need for further investigation into the interaction between diversity, structure, and task type. Future work should explore how to select or design persona combinations that are both diverse and cohesive.

\subsection{Evaluating Team Comprehension and Coordination}

\paragraph{Pre-task priors}
As outlined in \S\ref{interaction dynamics}, agents are queried before team interaction to gauge their priors. Q$^{\text{pre}}_1$ and Q$^{\text{pre}}_2$ assess agents' understanding of the shared team goal and their individual roles. We compute log odds~\cite{barnard_statistical_2018} to identify words disproportionately used across groups, revealing how agents internalize team objectives.

For Q$^{\text{pre}}_1$ top log-odds show that flat teams emphasize efficiency and coordination (e.g.,``wellorganized,'' ``guide,'' ``facilitate''), while hierarchical teams highlight structured, task-oriented language (e.g., ``brainstorming,'' ``development,'' ``provided''). Similarly for Q$^{\text{pre}}_2$, flat teams emphasize collective action and coordination (e.g., ``facilitate,'' ``collective,'' ``wellorganized''), while hierarchical teams reference structured processes and delegation (e.g., ``provided,'' ``decisionmaking,'' ``wellstructured'').  

When assessing effect of diversity, in both Q$^{\text{pre}}_1$ and Q$^{\text{pre}}_2$,  we observe that flat teams reference ``members'' and ``finding,'' suggesting greater awareness of group dynamics, whereas hierarchical teams remain consistent. 
These lexical patterns reflect how both team structure and demographic framing influence how agents conceptualize their roles and team objectives. A complete list of top log-odds words is provided in Table~\ref{tab:log_odds_q1} and Table~\ref{tab:log_odds_q2}  in the Appendix~\ref{appendix section: log-odds}.



We analyze average scores for Q$^{\text{pre}}_3$ to Q$^{\text{pre}}_5$, which assess agents' confidence in self, confidence in team, and expected team comprehension, across different team settings. Hierarchical teams report greater confidence in their team's ability to perform and integrate, whereas flat teams exhibit higher individual confidence. 
Team diversity has minimal effect on perceived self or team efficacy in flat teams, but it leads to a decline in overall confidence when introduced in hierarchical teams. For both, flat and hierarchical teams, when examined by level of diversity, we find that highly diverse teams tend to show greater individual confidence, while low-diversity teams express more confidence in team comprehension. These findings suggest that both team structure and diversity shape how agents anticipate their collaborative dynamics before task execution. Additional detailed results are provided in 
Table~\ref{tab:pretask_ttest}-\ref{tab:pre_high_low} in the Appendix~\ref{appendix section: detailed analysis}.

\begin{figure}
    \centering
    \includegraphics[width=75mm]{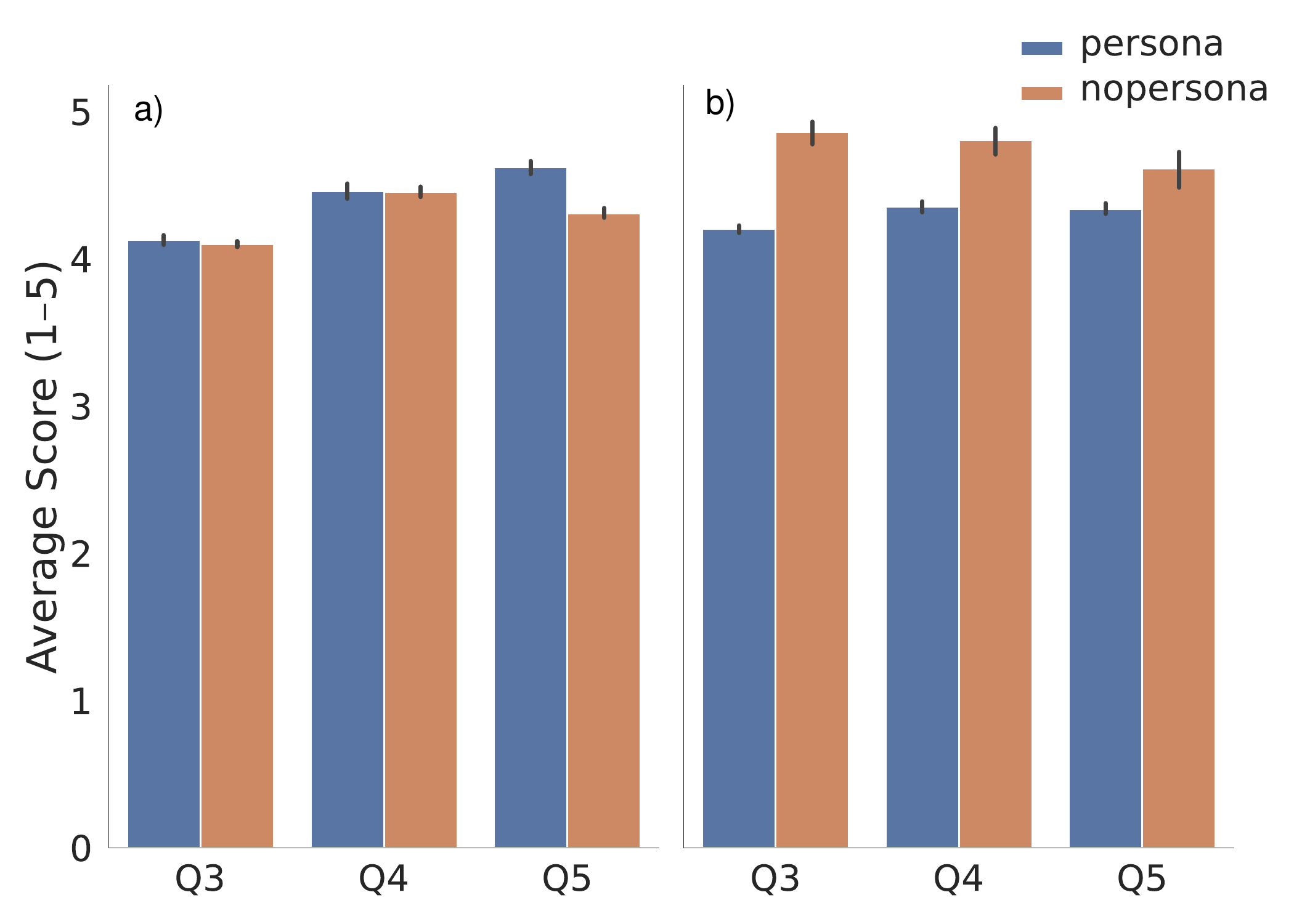}
    \caption{Average score for Q$^{\text{pre}}_3$, Q$^{\text{pre}}_4$, Q$^{\text{pre}}_5$. a) flat structure. b) hierarchical structure.}
    \label{fig:preint_personascores}
\end{figure} 

\begin{figure}
    \centering
    \includegraphics[width=1\linewidth]{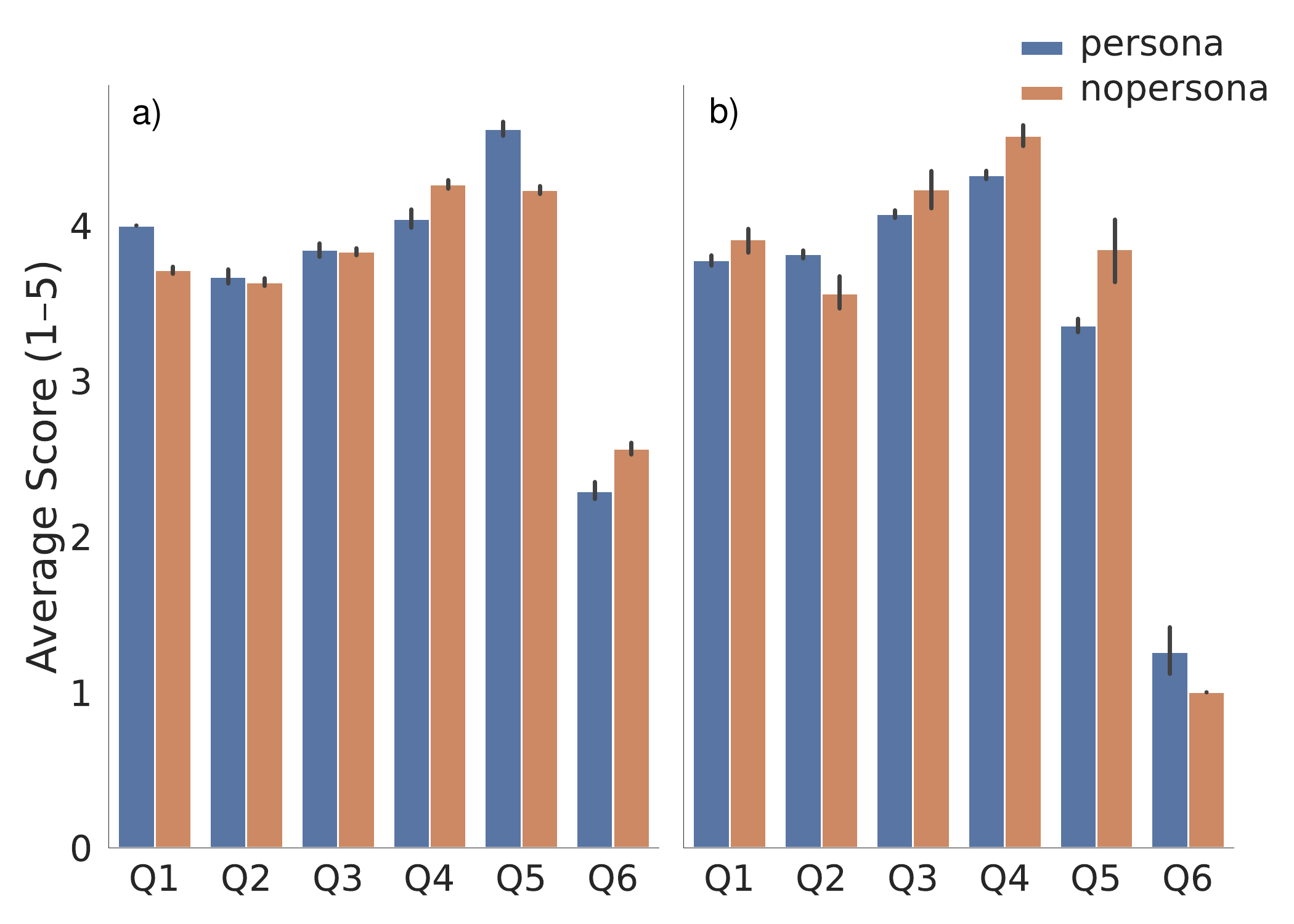}
    \caption{Average score all post-elicitation probing. a) flat structure. b) hierarchical structure.}
    \label{fig:postint_scores_persona}
\end{figure}

\paragraph{Post-task shifts}
In the post-task elicitation-probing analysis, we observe distinct patterns across team structures and diversity conditions. Figure~\ref{fig:postint_scores_persona} shows the post-task elicitation-probing scores across all agents in all team settings. We map each confidence-related pre-task item to its post-task counterpart (Q$^{\text{pre}}_3$ $\leftrightarrow$ Q$^{\text{post}}_2$, Q$^{\text{pre}}_4$ $\leftrightarrow$ Q$^{\text{post}}_3$, Q$^{\text{pre}}_5$ $\leftrightarrow$ Q$^{\text{post}}_4$)  to assess change in perceived collaboration over time. 
Each pair highlight the observed differences in perceived individual contribution, team contribution and team comprehension. We find that in all team settings, pre-elicitation probing acquired higher scores, indicating that post-task, there is a decrease in the confidence and perception of agent and team performance, and team comprehension. 

Flat teams tend to foster a stronger sense of individual achievement, while hierarchical teams have a more positive perception of team contribution and comprehension. Diversity impairs coordination in hierarchical teams but enhances cohesion and perceived contribution in flat, decentralized structures. These findings point to nuanced effects of structure and diversity on how agents perceive their contributions and collective outcomes after collaboration. Additional detailed results are provided in 
Table~\ref{tab:post_flat_vs_hier}-\ref{tab:post_ttest_diversity_highlow} in the Appendix~\ref{appendix section: detailed analysis}.

\subsubsection{Evaluation of AI conversations}

\paragraph{Human evaluation} \label{human_eval}
To assess the reliability of GPT-4o as an evaluator for the AI team conversations, we collected expert annotations familiar with textual data and LLM outputs. Using a coverage-first sampling method, we randomly sampled at least three examples from both flat and hierarchical teams under all models and datasets, balancing diversity conditions.
This yielded us 200 examples—--sufficient for human annotation feasibility and statistical adequacy (95\% CI of 6.5\% for binomial outcomes).

In the first round, the expert annotators independently annotated 50 of the 200 sampled conversations, based on the questions provided in section ~\ref{interaction dynamics}. They then refined the interpretation and scoring criteria through discussion to improve consistency. The remaining 150 conversations were annotated independently, and Inter-Annotator Agreement (IAA) was computed. Annotators achieved a 33\% exact match and 58.3\% within a one-point difference. 
Each annotator must annotate each conversation on 5 subjective questions on a 5-point scale. The subjectivity and high-dimensionality of scoring, makes reliable standard IAA score such as Fleiss's Kappa or Krippendorf's Alpha challenging~\cite{amidei-etal-2018-rethinking, yoshino-etal-2018-dialogue, Uymaz2023Collaborative}. 

\paragraph{A few-shot prompt using  examples aligned with human experts}
To calibrate GPT-4o as an evaluator, we constructed a few-shot prompt including additional instructions that were provided to human annotators. Additionally, we sampled 12 conversations with average human annotator scores across the 5 questions, while maintaining the distribution of scores. To verify if GPT-4o matches the grounding, we calculate the correlation between average human annotations and GPT-4o scores. We find Spearman correlation on the 150 sampled conversations to be 0.42, which is deemed to be a fair correlation. The few short prompt was then scaled to a sample of 2,500 randomly selected conversations while ensuring uniform representation of various team settings. The results of this scaled GPT-4o annotation is reported below.

\paragraph{LLM-judged conversation quality}
GPT-4o evaluation, following the LLM-as-a-judge approach, suggests that flat teams outperform hierarchical teams across all dimensions. On average, flat teams receive higher scores in \textit{Team Comprehension} ($3.34$ vs.\ $2.73$), \textit{Collaboration} ($2.50$ vs.\ $2.42$), \textit{Coherence} ($3.02$ vs.\ $2.61$), \textit{Reasoning Strength} ($3.42$ vs.\ $2.65$), and \textit{Confidence in Final Answer} ($3.31$ vs.\ $2.58$).
This may be because flat teams exhibit more balanced reasoning and coordination, which align better with GPT-4o's evaluation preferences.
Notably, team diversity further boosts GPT-4o evaluations in flat teams across all metrics, while in hierarchical teams, teams with diversity show only marginal improvements—or even slightly lower ratings in some dimensions—indicating that diversity aids collaboration primarily in settings with open, peer-based communication. 
Additional detailed results are provided in 
Table~\ref{tab:gpt4o_structure_persona} in the Appendix~\ref{appendix section: GPT4o}.

\section{Conclusion}

Can lessons from human team science inform the design of multi-agent LLM systems? Our findings suggest that team structure, diversity, and interaction dynamics each play a critical role in shaping team outcomes.

Our findings mirror several well-established patterns from human team science. Decentralized (flat) structures perform better on tasks requiring exploration, creativity, or ambiguous reasoning, where parallel idea generation is advantageous~\cite{evans_flatteam_innovation}. Similarly, our LLM teams excel on commonsense reasoning tasks that lack clear decomposition. Flat teams consistently outperform hierarchical ones across reasoning tasks, particularly in multi-step problems where open communication enables more effective coordination.

Although agents report high pre-task confidence, post-task reflections reveal difficulties in integrating diverse perspectives—especially in hierarchical teams, where constrained communication limits mutual understanding. GPT-4o evaluations corroborate these struggles, showing that hierarchical teams perform worse overall.

The effects of team diversity are more nuanced. Consistent with human studies~\cite{Mannix_neale_2005}, diversity can enrich reasoning through varied perspectives but also introduces coordination friction when social alignment is weak. In our LLM teams, diversity often reduces accuracy but fosters more reflective, self-aware collaboration—a paradox also observed in human teams. These results suggest that diversity positively shapes team dynamics even when it does not improve task performance. Future work should examine how structural design can better support diversity-aware coordination to harness these benefits without sacrificing efficiency.

Future work should explore adaptive team structures that dynamically adjust roles, delegation, and communication patterns based on task complexity and team composition. Learning-based coordination strategies, such as reinforcement learning or meta-optimization, may help align structure with agent capabilities and diversity profiles.
Furthermore, to allow for a deeper comparison between LLM and human teams, it would be a valuable direction to build tasks and frameworks that more directly probe social influence, group identity, and conflict resolution. Within hierarchical teams, specifically, investigating power dynamics and influential power of the leader could be an interesting direction to take. 
Evaluating these approaches in multilingual, cross-cultural, and real-world tasks would further test the generalizability of our findings. Finally, further research is needed on interpretability and accountability in multi-agent systems, particularly in understanding how teams reason, disagree, and converge on decisions over time.

\clearpage

\section*{Limitations}

We highlight some key limitations of this study. 
First, our experiments are constrained to the English-language prompts and may not generalize to multi-lingual settings. 
Second, we operationalize diversity using demographic attributes, which serves only as surface-level proxies for deeper experiential and cultural variation. 
Third, we use relatively small open-source models (7–8B), which may constrain agents' interaction capabilities and reasoning depth. 
Fourth, our team design does not incorporate dynamic or adaptive strategies, which is a fundamental aspect of interactions. This may have exacerbated communication bottlenecks and misalignment between team members.
Fifth, we only use datsets that focus on natural language reasoning tasks from pre-2021 era. This possesses an issue of data leakage and additionally does not allow to explore all features or elements of team dynamics such as coordination and negotiation. This could imply that commonsense reasoning favors flat teams, as hierarchical teams are known to be more advantageous in more complex, multi-step tasks with clear subtasks, where role specialization and communication constraints could improve efficiency~\cite{Zhou2013Designing, Anicich2015Hierarchical, Wellman2020Beyond}. Exploring such tasks is a valuable direction for future work, and we see our current results as a necessary baseline to understand how different structures perform on less structured but cognitively foundational tasks.
Lastly, our post-task reflections and confidence measures offer only coarse-grained approximations of agent meta-cognition. Any interpretations of agent ``perception'' or ``awareness'' should thus be made cautiously.

These limitations point to  broader challenges in deploying multi-agent systems that simulate human-like teams. 
Naively implementing diversity or structure without sensitivity to coordination dynamics may result in degraded performance, tokenistic representation, or unintended social consequences. 
In high-stakes domains (e.g., education, healthcare, policy deliberation), misaligned agent teams may reinforce existing biases or produce  misleading outcomes under the appearance of deliberative reasoning.

\bibliography{custom}

\clearpage
\appendix

\section{Appendix}
\label{sec:appendix}
\subsection{Implementation Details} \label{appendix section: implementation}
We provide further implementation details of our study in this section. The temperature for all models---Llama-8b-Instruct, Qwen, Mistral, and Deepseek---are set to 0.7 as though are the defaults. We provide the algorithms of our workflows in Algorithm~\ref{alg:flat-team} and Algorithm~\ref{alg:hier-team}. For the evaluation of conversations by GPT-4o, we provided a temperature of 0.7. The final output for both workflow is one of the multiple options provided for the question. For testing our framework, we use A800 GPUs with 80GB. 

\subsection{Additional Justification}


\begin{algorithm}[ht]
\small
\caption{\textsc{FlatTeamDebate}: Multi-Round Discuss-and-Vote Framework}
\label{alg:flat-team}

\KwIn{Question $Q$; maximum rounds $R$; agents $\mathcal{A} = \{A_i\}_{i=1}^{n}$ (odd $n$)\;}
\KwOut{Team answer $\hat{a}$\;}

$r \leftarrow 0$\;

\While{$r \le R$ \textbf{and} \textsc{Consensus}($\{a_i^{(r-1)}\}$) = \texttt{false}}{
  \ForEach{agent $A_i \in \mathcal{A}$}{
    \uIf{$r = 0$}{
      $P \leftarrow Q$\;
    }
    \Else{
      $P \leftarrow (Q,\ \{a_j^{(r-1)}\}_{j=1}^{n})$\;
    }
    $(a_i^{(r)}, e_i^{(r)}, p_i^{(r)}) \leftarrow A_i(P)$\;
  }
  $r \leftarrow r + 1$\;
}
$\hat{a} \leftarrow$ \textsc{MajorityVote}($\{a_i^{(r-1)}\}$)\;
\Return $\hat{a}$\;
\end{algorithm}

We use the following prompts for Flat Team. For tests with diversity, we provide the demographic diversity of agents in each round. 
\begin{tcolorbox}[title=Prompt Design for Flat Team Structure, 
                  breakable, 
                  colback=blue!3!white, 
                  colframe=blue!12!white,
                  fonttitle=\bfseries,
                  coltitle=black,
                  sharp corners=south,
                  boxrule=0.5pt,
                  arc=2pt,
                  enhanced,
                  width=\linewidth]
\small

\textbf{Round 0 (Initial Answer)}\\
\textit{You are a reasoning agent \texttt{{agent\_id}}. You are here to answer multiple choice reasoning questions.}

\textit{Please answer the following question by selecting only one option.}

\textbf{Question:} \texttt{{q}}\\
\textbf{Answer:} \_\_\_

\vspace{0.8em}
\textbf{Rounds 1 to N (Refinement Phase)}\\
\textit{You are a reasoning agent \texttt{{agent\_id}}. You are here to answer multiple choice reasoning questions.}

\textit{You are part of a team of agents. You are expected to help your team get to the correct answer.}

\textit{You will be given the question and your previous response and your team members’ previous responses.}

\textit{Here are your previous answers from your team: \texttt{{context}}}

\textit{Take a moment to reflect on the responses and then engage in conversation to come to the right answer.}

\textbf{Question:} \texttt{{q}}\\
\textbf{Answer:} \_\_\_

\vspace{0.8em}
\textbf{Final Round (Consensus Prompt)}\\
\textit{Agents, review the conversation: \texttt{{final\_context}}.}

\textit{Come to a consensus on the best final answer for the question:}

\textbf{Question:} \texttt{{q}}\\
\textbf{Answer:} \_\_\_

\end{tcolorbox}

\begin{algorithm}[ht]
\small
\caption{\textsc{HierTeam}: Leader–Subordinate Delegation Framework}
\label{alg:hier-team}

\KwIn{Question $Q$; maximum rounds $R$; leader $L$; subordinates $\mathcal{S} = \{S_k\}_{k=1}^{m}$\;}
\KwOut{Final team answer $\hat{a}$\;}

$r \leftarrow 0$\;

\While{$r \le R$}{
  \uIf{$r = 0$}{
    $P_L \leftarrow Q$\;
  }
  \Else{
    $P_L \leftarrow (Q,\ \{a_k^{(r-1)}\}_{k=1}^{m})$\;
  }

  $\{I_k^{(r)}\}_{k=1}^{m} \leftarrow L(P_L)$ \tcp*{Leader generates instructions}

  \ForEach{$S_k \in \mathcal{S}$}{
    $(a_k^{(r)}, e_k^{(r)}) \leftarrow S_k(I_k^{(r)})$\;
  }

  \If{$r = R$}{
    $\hat{a} \leftarrow L(\{a_k^{(r)}\}_{k=1}^{m})$\;
    \Return $\hat{a}$\;
  }

  $r \leftarrow r + 1$\;
}
\end{algorithm}

Similarly, based on Algorithm~\ref{alg:hier-team}, we created the following prompt flow for testing hierarhical teams. 

\begin{tcolorbox}[title=Prompt Design for Hierarchical Team Structure, 
                  breakable, 
                  colback=purple!3!white, 
                  colframe=purple!80!black,
                  fonttitle=\bfseries,
                  coltitle=black,
                  sharp corners=south,
                  boxrule=0.5pt,
                  arc=2pt,
                  enhanced,
                  width=\linewidth]
\small

\textbf{Round 0 – Leader's Initial Instruction Prompt}\\
\textit{You are the team leader of a reasoning team. The goal of the team is to answer reasoning questions as accurately as possible. You manage the following agents: \texttt{{team\_description}}.}

\textit{Your role is to delegate tasks to your team members so that they can provide you with useful information.}

\textit{Create clear instructions for each agent.}

\textbf{Question:} \texttt{{q}}\\
\textbf{Team Members:} \texttt{{team\_members}}\\
\textbf{Instructions:}\\
\textit{Agent 1: \_\_\_}\\
\textit{Agent 2: \_\_\_}\\
\textit{Agent 3: \_\_\_}

\vspace{1em}
\textbf{Round 1 – Team Member Reasoning Prompt}\\
\textit{You are a team member of a reasoning team. \texttt{{persona}} You are led by team leader Agent 1. Your role is to answer based on the leader's instruction to help solve the reasoning question.}

\textbf{Question:} \texttt{{q}}\\
\textbf{Instruction:} \texttt{{instr}}\\
\textbf{Answer:} \_\_\_

\vspace{1em}
\textbf{Round 1 to N – Leader's Refinement Instruction Prompt}\\
\textit{You are the team leader of a reasoning team. You manage the following agents: \texttt{{team\_description}}. Your team members have submitted initial answers to the question.}

\textit{Review their responses and provide each member with updated instructions to reaffirm or correct their reasoning.}

\textit{Your instructions must be clear and under 10 words.}

\textbf{Team Members’ Answers:} \texttt{{team\_answers}}\\
\textbf{Question:} \texttt{{q}}\\
\textbf{Instructions:}\\
\textit{Agent 1: \_\_\_}\\
\textit{Agent 2: \_\_\_}\\
\textit{Agent 3: \_\_\_}

\vspace{1em}
\textbf{Final Round – Leader Final Reflection Prompt}\\
\textit{You are the team leader of a reasoning team. Your team members have responded based on your updated instructions.}

\textit{Reflect on their responses and provide the final correct answer. Your answer may differ from your team members'.}

\textbf{Team Members’ Final Answers:} \texttt{{team\_answers}}\\
\textbf{Question:} \texttt{{q}}\\
\textbf{Final Answer:} \_\_\_

\end{tcolorbox}

\subsection{Licenses}
We document the language, domain, and demographic characteristics of the datasets and models used in this study following best practices from data and model documentation toolkits (e.g., Data Statements, Model Cards, Datasheets for Datasets). All datasets—CommonsenseQA, Social IQa, StrategyQA, and Implicit Hate Detection—are in English and focus on reasoning tasks across different domains. CommonsenseQA and StrategyQA cover general knowledge and commonsense reasoning, while Social IQa focuses on social and situational commonsense, and Implicit Hate captures nuanced, often sarcastic, toxic speech from Reddit. Demographic information about dataset authors is generally unavailable, with the exception of Social IQa and Implicit Hate, which were crowd-annotated by U.S.-based workers with diverse backgrounds. The models we use include Meta’s LLaMA-8B Instruct, Alibaba’s Qwen-7.5B Instruct, Mistral-7B v0.3 Instruct, and DeepSeek R1 (a distilled LLaMA variant). All models are primarily trained on English, with partial multilingual capabilities in Qwen and Mistral. Their training data spans web text, code, and instruction-tuned corpora, although precise data composition is not fully disclosed for all models. None of the models guarantee demographic balancing or fairness-aware pretraining. Model licenses range from fully permissive (Apache 2.0, MIT) to research-constrained (LLaMA 2 Community License).

\begin{table}[ht]
\centering
\caption{Licenses for datasets used in this study.}
\label{tab:dataset_licenses}
\begin{tabular}{llp{7cm}}
\toprule
\textbf{Dataset} & \textbf{License}  \\
\midrule
CommonsenseQA    & CC BY-SA 4.0  \\
Social IQa       & MIT License  \\
StrategyQA       & Apache 2.0  \\
Implicit Hate    & MIT License  \\
\bottomrule
\end{tabular}
\end{table}

\begin{table*}[ht]
\centering
\caption{Licenses for models used in this study.}
\label{tab:model_licenses}
\begin{tabular}{lll}
\toprule
\textbf{Model} & \textbf{Organization} & \textbf{License} \\
\midrule
LLaMA-8B Instruct            & Meta         & LLaMA 2 Community License \\
Qwen-2.5-7B Instruct           & Alibaba      & Apache 2.0 \\
Mistral-7B Instruct v0.3     & Mistral      & Apache 2.0 \\
DeepSeek R1 (LLaMA-8B Distil)& DeepSeek     & MIT License \\
\bottomrule
\end{tabular}
\end{table*}

\subsection{Diversity Settings} \label{diversity_setting}
Demographic diversity of an agent is constructed using four dimensions of human demographics: age, gender, ethnicity, and occupation. Each dimension includes a range of categories—--gender (male, female), age (young, young working professional, working professional, senior citizen), ethnicity (White, Asian, Black), and occupation (white-collar, blue-collar). By systematically combining these categories, we generate a total of 48 unique personas used to populate our teams. 
Here is an example of the persona provided to an agent: 
\begin{quote}
    \textit{You are male and of age 18 to 24. You identify as white and work a blue collar job.}
\end{quote}

\section{Additional Results} \label{Appendix_sec: Additional_results}
\subsection{Single Agent}

To understand the effect of persona, we conducted an ablation study using single agents. We test the effect of the four dimensions of diversity as mentioned in Section~\ref{team diversity} on the CS and SQA datasets. We examine the effect of each dimension and further the combination of these dimensions. Table~\ref{tab:single_agent_dimension_test} shows the average performance delta of diversity-based agent compared to their no-diversity counterparts, grouped by the number of demographic dimensions used. When only one or two dimensions are included (e.g., just gender or age), teams see modest performance improvements, with an average delta of $1.83$ and $0.75$, respectively. However, as the number of persona dimensions increases to three or four, performance begins to decline. Teams with four-dimensional personas (gender, age, ethnicity, occupation) show a negative average delta of $-0.91$, suggesting potential cognitive overload or misalignment introduced by more complex social cues. These results indicate that while lightweight demographic cues may support collaboration, higher-dimensional personas may hinder team effectiveness, possibly due to increased coordination demands or difficulty in integrating diverse perspectives.

\begin{table}[ht]
\small
\centering
\caption{Effect of Persona Dimensionality on Single-Agent Accuracy (Delta from No-Persona Baseline)}
\begin{tabular}{cccc}
\toprule
\textbf{\# Dimensions} & \textbf{Avg Accuracy Delta} & \textbf{Std Dev} \\
\midrule
1 & +1.83 & 2.05  \\
2 & +0.75 & 3.53 \\
3 & $-1.62$ & 8.70  \\
4 & $-0.91$ & 9.71  \\
\bottomrule
\end{tabular}
\label{tab:single_agent_dimension_test}
\end{table}

\subsection{Comparing different sizes and rounds}
As mentioned in Section~\ref{team structure}, we create teams of varying size. In flat teams, we test across teams size of 3, 5 and 7 for 2 to 4 rounds. In Table~\ref{tab:accuracy_flatteam_setting}, we report the average bootstrapped accuracy of all flat team settings across the four datasets. We find that the number of rounds and number of agents have very small effect on the performance of the team. These findings indicate that scaling in teams is a nuanced issue that future work can address in the context of team science for AI teams

\begin{table}[ht]
\small
\centering
\caption{Average accuracy (\%) across tasks by team setting (agents $\times$ rounds).}
\begin{tabular}{lcccc}
\toprule
\textbf{Team Setting} & \textbf{CS} & \textbf{IH} & \textbf{SQA} & \textbf{ST} \\
\midrule
3 agents, 2 rounds & 71.29 & 43.43 & 54.41 & 63.97 \\
3 agents, 3 rounds & 71.98 & 38.22 & 55.11 & 63.49 \\
3 agents, 4 rounds & 71.94 & 41.47 & 54.65 & 63.68 \\
5 agents, 2 rounds & 72.30 & 38.63 & 54.08 & 64.51 \\
5 agents, 3 rounds & 73.45 & 37.56 & 54.25 & 64.91 \\
5 agents, 4 rounds & 72.30 & 37.69 & 54.16 & 64.34 \\
7 agents, 2 rounds & 67.98 & 42.06 & 55.17 & 64.84 \\
7 agents, 3 rounds & 69.28 & 42.16 & 54.52 & 65.00 \\
7 agents, 4 rounds & 68.26 & 42.52 & 54.89 & 64.33 \\
\bottomrule
\end{tabular}
\label{tab:accuracy_flatteam_setting}
\end{table}

As mentioned in Section~\ref{team structure}, we create two hierarchical teams, one with 1 leader and 3 team members and second with 1 leader, 2 managers and 4 team members. The team settings emulate 1 level and 2 levels of hierarchy respectively. In the paired t-test comparing the performance of these two settings, we find that 1 level of hierarchy is consistently preferred, potentially indicating that more rigid structures of communications are not beneficial. 

\begin{table}[ht]
\centering
\small
\caption{Paired t-test comparing hierarchical level 1 vs. hierarchical level 2 team performance across tasks.}
\begin{tabular}{lrrrr}
\toprule
\textbf{Task} & \textbf{t-stat} & \textbf{Mean Difference} & \textbf{Cohen's d} \\
\midrule
CS & 1.978  & 21.561 & 0.989 \\
IH  & 2.873$^{\ast}$ & 7.713  & 1.437 \\
SQA & 5.280$^{\ast}$ & 19.798 & 2.640 \\
ST  & 3.125$^{\ast}$ & 21.917 & 1.563 \\
\bottomrule
\end{tabular}
\end{table}


Table~\ref{tab:team_structure_tstat} reports $t$-test statistics comparing team performance under diversity versus no-diversity conditions across a variety of team configurations. Results indicate that diversity often has a statistically significant effect, but the direction and magnitude vary by task and team size. For example, in larger teams (5–7 agents), diversity has more pronounced positive effects on ST, particularly as team size and number of rounds increase. Notably, the negative impact of diversity on CS is consistent across all team sizes, suggesting that uniformity in perspective may benefit certain types of commonsense reasoning. Overall, the data underscores the nuanced and configuration-dependent impact of diversity on team reasoning dynamics.

\begin{table}[ht]
\centering
\small
\caption{Paired $t$-test statistics for diversity vs. no-diversity comparisons across team settings. Asterisk ($^*$) indicates $p < 0.05$.}
\begin{tabular}{lcccc}
\toprule
\textbf{Team Setting} & \textbf{CS} & \textbf{IH} & \textbf{SQA} & \textbf{ST} \\
\midrule
3 agents, 2 rounds & -9.63$^*$ & 3.01$^*$ & -2.42$^*$ & 0.91 \\
3 agents, 3 rounds & -7.30$^*$ & 0.25     &  0.30     & -5.54$^*$ \\
3 agents, 4 rounds & -7.12$^*$ & 7.15$^*$ & -1.89     & -7.77$^*$ \\
5 agents, 2 rounds & -9.68$^*$ & -2.42$^*$ & -3.71$^*$ & 9.02$^*$ \\
5 agents, 3 rounds & -10.15$^*$ & -2.56$^*$ & -4.38$^*$ & 5.79$^*$ \\
5 agents, 4 rounds & -10.28$^*$ & -2.50$^*$ & -4.33$^*$ & 10.39$^*$ \\
7 agents, 2 rounds & -6.04$^*$ & 2.60$^*$  & -1.15     & 9.12$^*$ \\
7 agents, 3 rounds & -5.64$^*$ & 2.76$^*$  & -3.12$^*$ & 6.07$^*$ \\
7 agents, 4 rounds & -5.68$^*$ & 1.38      & -3.20$^*$ & 8.69$^*$ \\
\bottomrule
\end{tabular}
\label{tab:team_structure_tstat}
\end{table}

Table~\ref{tab:ttests_tasks} presents paired $t$-test statistics comparing diversity-based teams to no-diversity teams across different hierarchical structures and reasoning rounds. The results show that diversity significantly enhances performance across all tasks in the first-level hierarchy (Hier. l1), with extremely high $t$-values and $p < .001$ for SQA and ST. However, the effects diminish or even reverse in the second-level hierarchy (Hier. l2), where deeper delegation and communication layers appear to hinder the effective use of persona information. Specifically, SQA shows a dramatic drop from strong positive gains in Hier. l1 to significant negative effects in Hier. l2, suggesting that tasks requiring nuanced social inference are particularly sensitive to how persona information is coordinated across levels. This pattern highlights that persona benefits are maximized in shallow hierarchies where agents can directly interpret and leverage identity cues.

\begin{table}[ht]
\small
\centering
\caption{T-test statistics (t-value) for persona vs. no-persona comparisons across tasks and team settings (Team). l1 refers to hierarchical teams with 1 leader and 3 team members, l2 refers to hierarchical teams with 1 leader, 2 managers and 4 team members. Significance is denoted as: * $p<.05$, ** $p<.01$, *** $p<.001$}
\label{tab:ttests_tasks}
\begin{tabular}{lcccc}
\toprule
\textbf{Team} & \textbf{CS} & \textbf{IH} & \textbf{SQA} & \textbf{ST} \\
\midrule
l1, 2 r. & 7.49***  & 7.09***  & 14.46*** & 17.38*** \\
l1, 3 r.  & 8.87***  & 7.31***  & 13.96*** & 19.57*** \\
l1, 4 r.  & 8.07***  & 4.61***  & 13.21*** & 17.90*** \\
l2, 2 r.  & 3.32**   & 0.20     & -6.25*** & 4.18***  \\
l2, 3 r. & 2.12*    & -0.37    & -6.43*** & 3.23**   \\
l2, 4 r.  & 1.22     & -0.30    & -5.49*** & 1.68     \\
\bottomrule
\end{tabular}
\end{table}

\subsection{Outperforming diversity teams} \label{App section: Outperforming}

Table~\ref{tab:outperforming} presents a comprehensive list of diversity-based team experiments that outperformed their no-diversity counterparts across various tasks, team sizes, and team structures. The results highlight the consistent advantage of persona use in flat teams, particularly those composed of 3 agents. In this configuration, CS and SQA dataset show strong gains, with experiments such as 04, 06, 07, 09, 11, and 14 repeatedly emerging across tasks and rounds. This suggests that certain team compositions—defined by the demographic personas assigned—are especially synergistic under conditions of open interaction and shared responsibility.

As team size increases to 5 or 7 agents, persona benefits remain evident, though the pattern becomes more task-specific. For instance, ST shows robust gains in larger flat teams, whereas gains for IH are sparse across all configurations. Hierarchical teams show more mixed results. While several experiments still outperform the baseline, especially in the first-level hierarchy (l1), the benefits of persona cues appear attenuated. In second-level hierarchies (l2), where communication is more constrained, persona advantage persists in SQA and ST but becomes less discriminative across configurations, likely due to reduced opportunities for mutual interpretation and integration.

These findings reinforce our broader claim that diversity does not uniformly improve performance, but rather is dependent on team structure and task. 

\begin{table*}[ht]
\small
\centering
\caption{Experiments where diversity-based teams outperformed their no-diversity counterparts.}
\label{tab:outperforming}
\begin{tabular}{>{\raggedright\arraybackslash}p{2.2cm} 
                >{\raggedright\arraybackslash}p{3 cm} 
                >{\raggedright\arraybackslash}p{3 cm} 
                >{\raggedright\arraybackslash}p{3 cm} 
                >{\raggedright\arraybackslash}p{3 cm}}
\toprule
\textbf{Team} & \textbf{CS} & \textbf{IH} & \textbf{SQA} & \textbf{ST} \\
\midrule
Flat (3 agents) & [01, 04, 05, 06, 07, 09, 11, 14] & [01, 07] & [01, 04, 06, 07, 10, 11, 12, 14, 15] & [] \\
Flat (5 agents) & [01, 03, 06, 08, 09, 10] & [] & [10, 11, 12] & [01, 02, 03, 04, 05, 06, 07, 08, 09, 10, 11, 12, 13] \\
Flat (7 agents) & [01, 03, 06, 08, 09, 10] & [] & [10, 11, 12] & [01, 02, 03, 04, 05, 06, 07, 09, 10, 12, 13] \\
Hier. l1 & [02, 03, 05, 06, 08, 10, 12, 13, 14] & [03, 06, 14] & [01, 02, 03, 04, 05, 06, 08, 09, 10, 12, 14] & [02, 03, 05, 08, 12, 13, 14] \\
Hier. l2 & [01, 03, 04, 09, 10, 11, 13] & [01, 03, 05, 06, 08, 10, 11, 13] & [01, 02, 03, 04, 05, 06, 07, 08, 09, 10, 11, 12, 13, 14] & [01, 02, 03, 04, 05, 06, 07, 08, 09, 10, 11, 12, 13, 14] \\

\bottomrule
\end{tabular}
\end{table*}

\subsection{Team Diversity} \label{appendix section: team div}

\begin{figure}
    \centering
    \includegraphics[width=\linewidth]{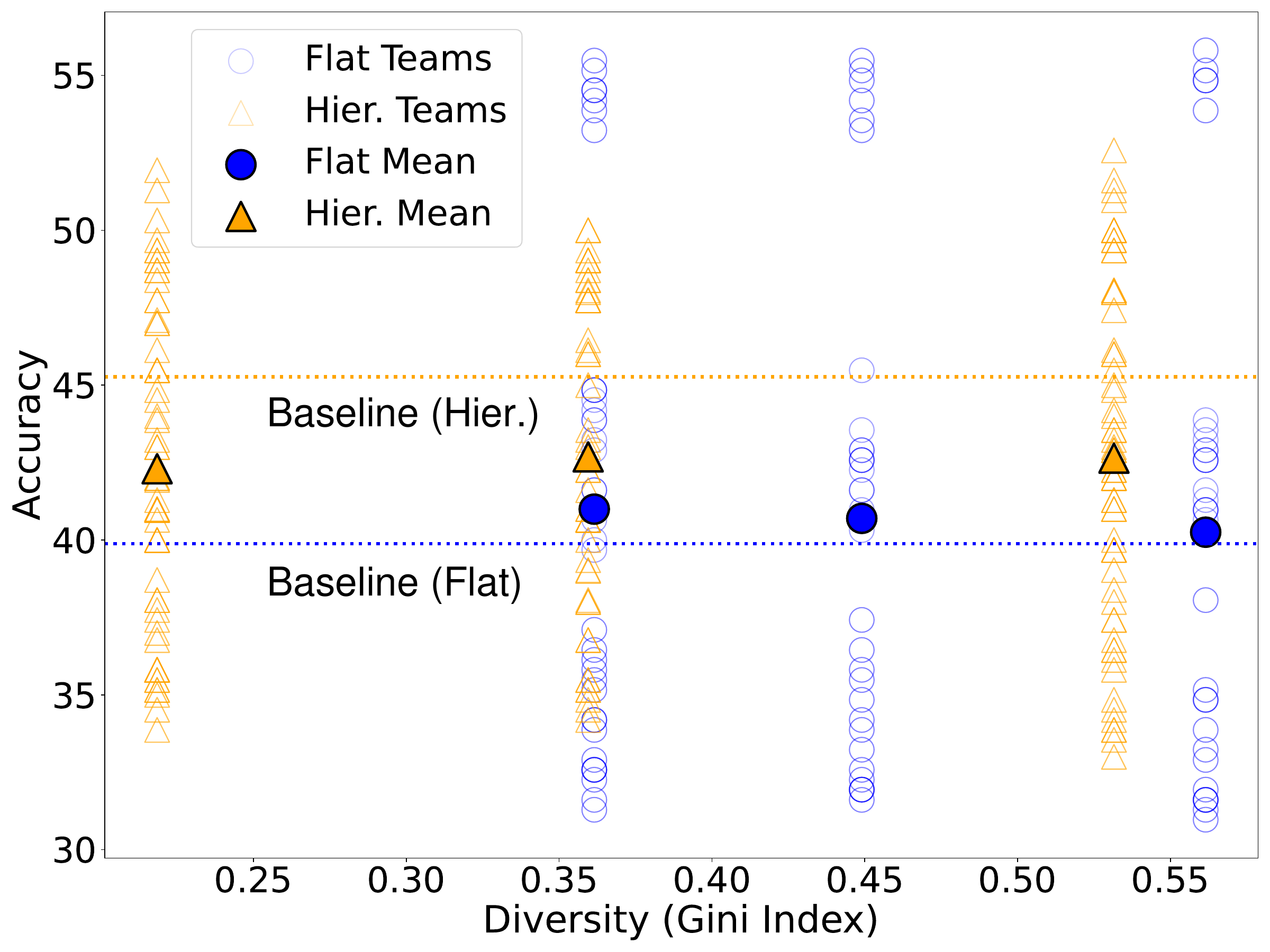}
    \caption{Trend of team diversity and performance in flat teams and hierarchical teams for Implicit Hate.} 
    
    \label{fig:diversity_performance_ih}
\end{figure}

\begin{figure}
    \centering
    \includegraphics[width=\linewidth]{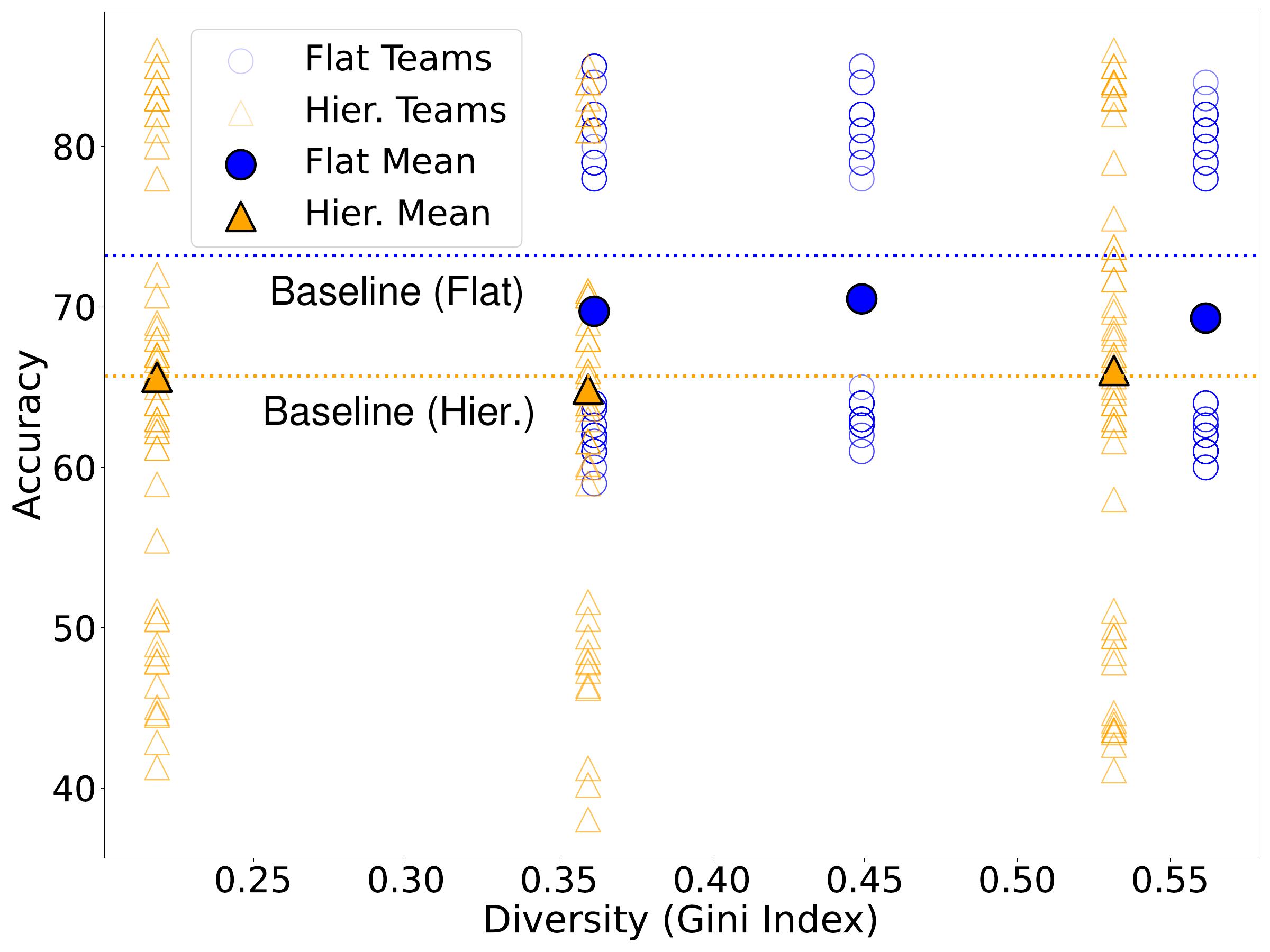}
    \caption{Trend of team diversity and performance in flat teams and hierarchical teams for CS dataset. $x$-axis represents the level of team diversity, calculated through Gini Index, and $y$-axis represents  performance of teams.}
    \label{fig:diversity_performance_cs}
\end{figure}

\begin{figure}
    \centering
    \includegraphics[width=\linewidth]{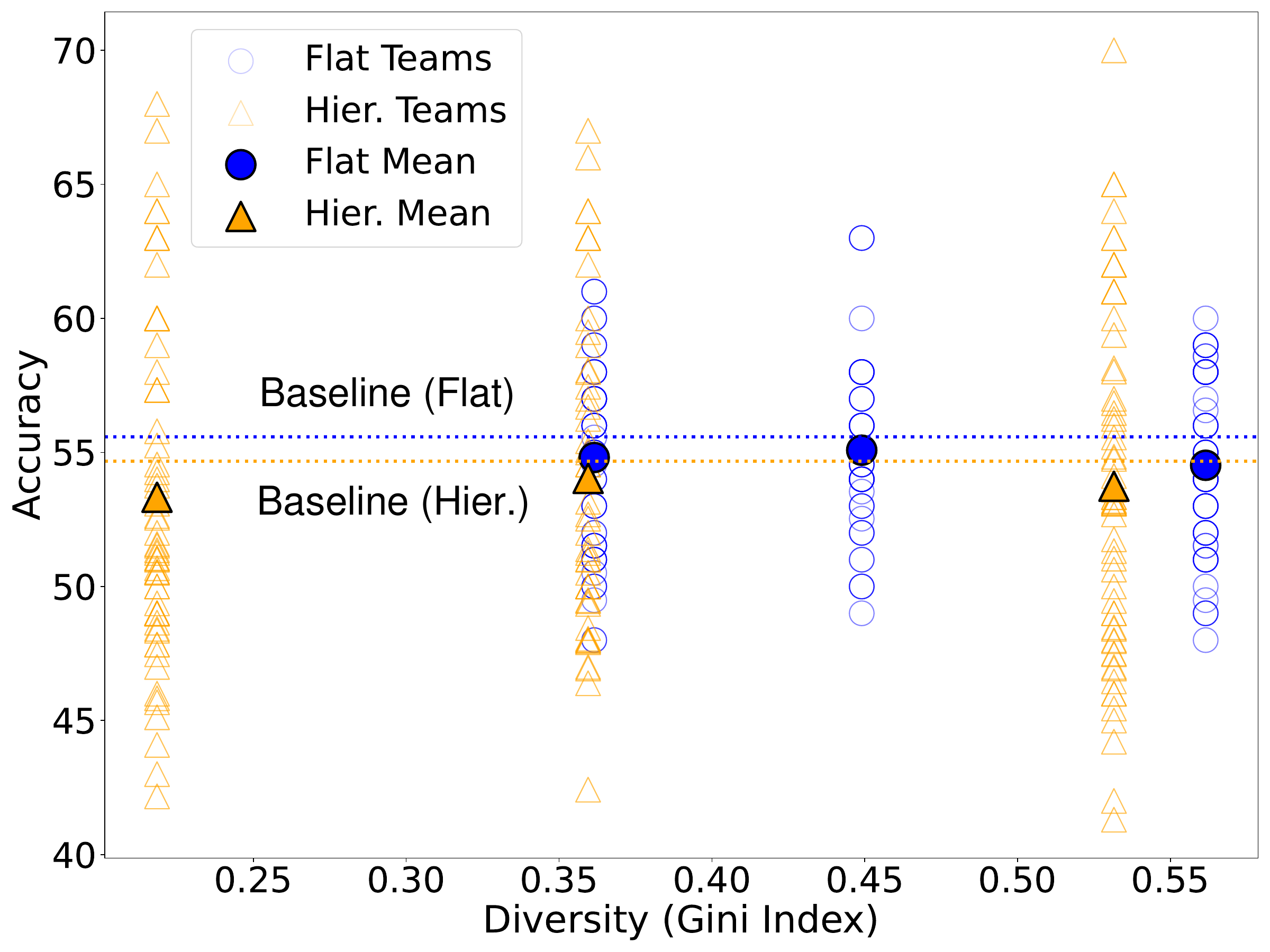}
    \caption{Trend of team diversity and performance in flat teams and hierarchical teams for SQA dataset. $x$-axis represents the level of team diversity, calculated through Gini Index, and $y$-axis represents  performance of teams.}
    \label{fig:diversity_performance_iqa}
\end{figure}

\begin{figure}
    \centering
    \includegraphics[width=\linewidth]{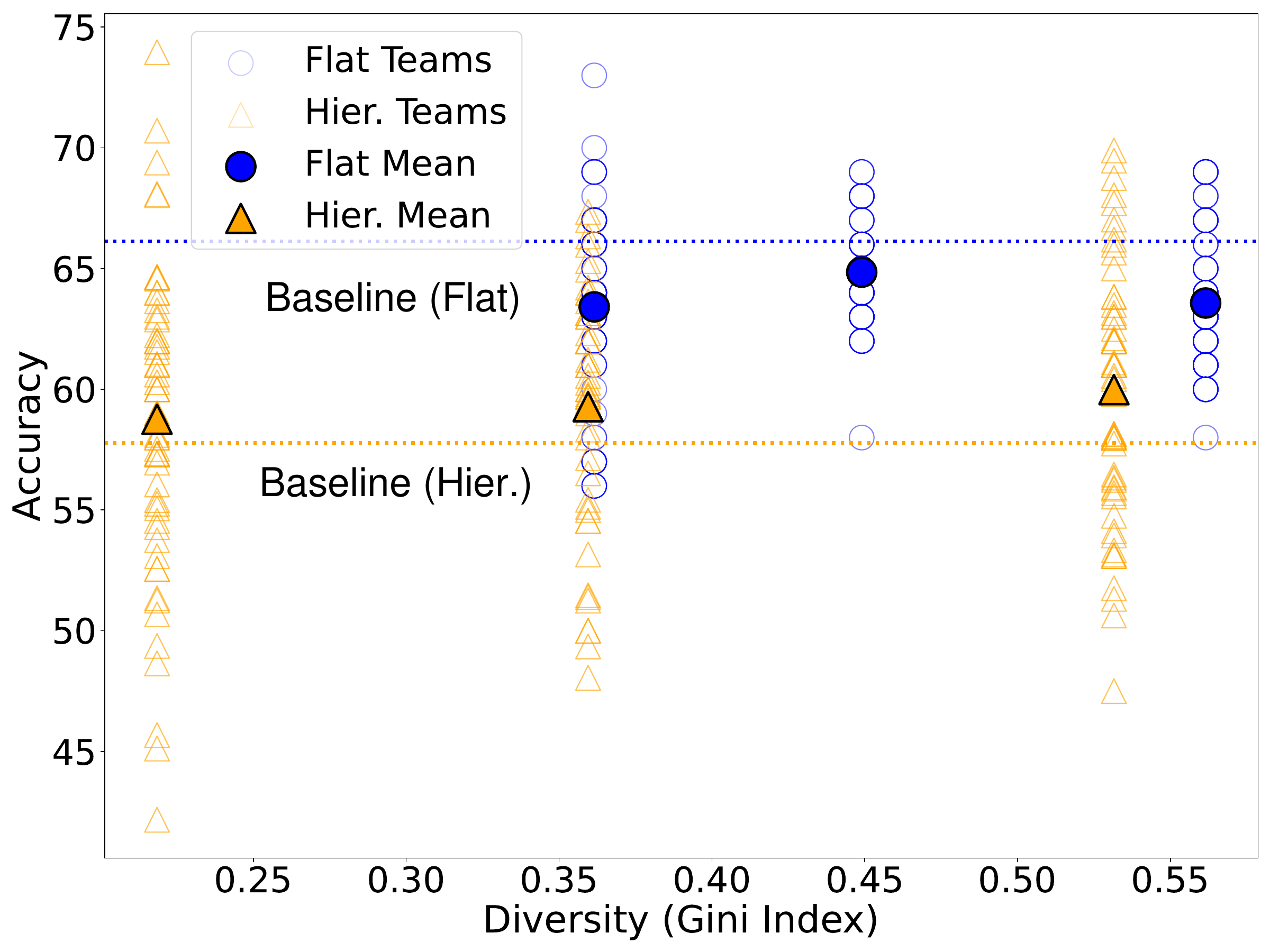}
    \caption{Trend of team diversity and performance in flat teams and hierarchical teams for ST dataset. $x$-axis represents the level of team diversity, calculated through Gini Index, and $y$-axis represents  performance of teams.}
    \label{fig:diversity_performance_st}
\end{figure}

Figures~\ref{fig:diversity_performance_cs},~\ref{fig:diversity_performance_iqa}, ~\ref{fig:diversity_performance_st},~\ref{fig:diversity_performance_ih} showcase the relationship between diversity, measured through Gini Index, against performance. Each figure shows the trend of flat and hierarchical teams with increasing diversity. We observe the trend the teams is dependent on the task. However, all three plots show high variance across the diversity teams, reaffirming that diversity does not have a universal effect on performance of teams. 

\subsection{Team Elicitation Probing}
\subsubsection{Pre-Task Prior Probing}\label{pre-task}

As mention in Section~\ref{interaction dynamics}, agents are queried before engaging in team interaction to gauge the agent prior. 
Each agent is asked to answer the following questions before the task. Q$^{\text{pre}}_1$ and Q$^{\text{pre}}_2$ are open-ended, while Q$^{\text{pre}}_3$–Q$^{\text{pre}}_5$ use a 1–5 scale (5 = highest):
\begin{enumerate}[label=Q$^{\text{pre}}_\arabic*.$, labelindent=0pt, leftmargin=*, noitemsep, topsep=0pt]
    \item What do you think is the primary goal of the team?
    \item What is your role in the team?
     \item How confident are you about executing the role?
    \item How confident are you in your team executing the task?
    \item How confident are you in the team’s ability to integrate diverse perspectives during the task?
\end{enumerate}

\subsubsection{Post-Task Reflection} \label{post-task}

After the completion of the task, post-elicitation querying is conducted to assess the observed shifts in agents. Agents respond to the following on a 1-5 scale (5 = highest):
\begin{enumerate}[label=Q$^{\text{post}}_\arabic*.$, labelindent=0pt, leftmargin=*, noitemsep, topsep=0pt]
\item How do you think your team performed to achieve the goal?
\item How well do you think you contributed to the team?
\item How well do you think your team members contributed to the team?
\item Were you able to understand your team members?
\item Do you think your team members understood you?
\item Do you think you could come up with these solutions that the group came with?
\end{enumerate}

\subsubsection{Wordclouds}

\begin{figure*}[t]
    \centering
    \includegraphics[width=1.1\linewidth]{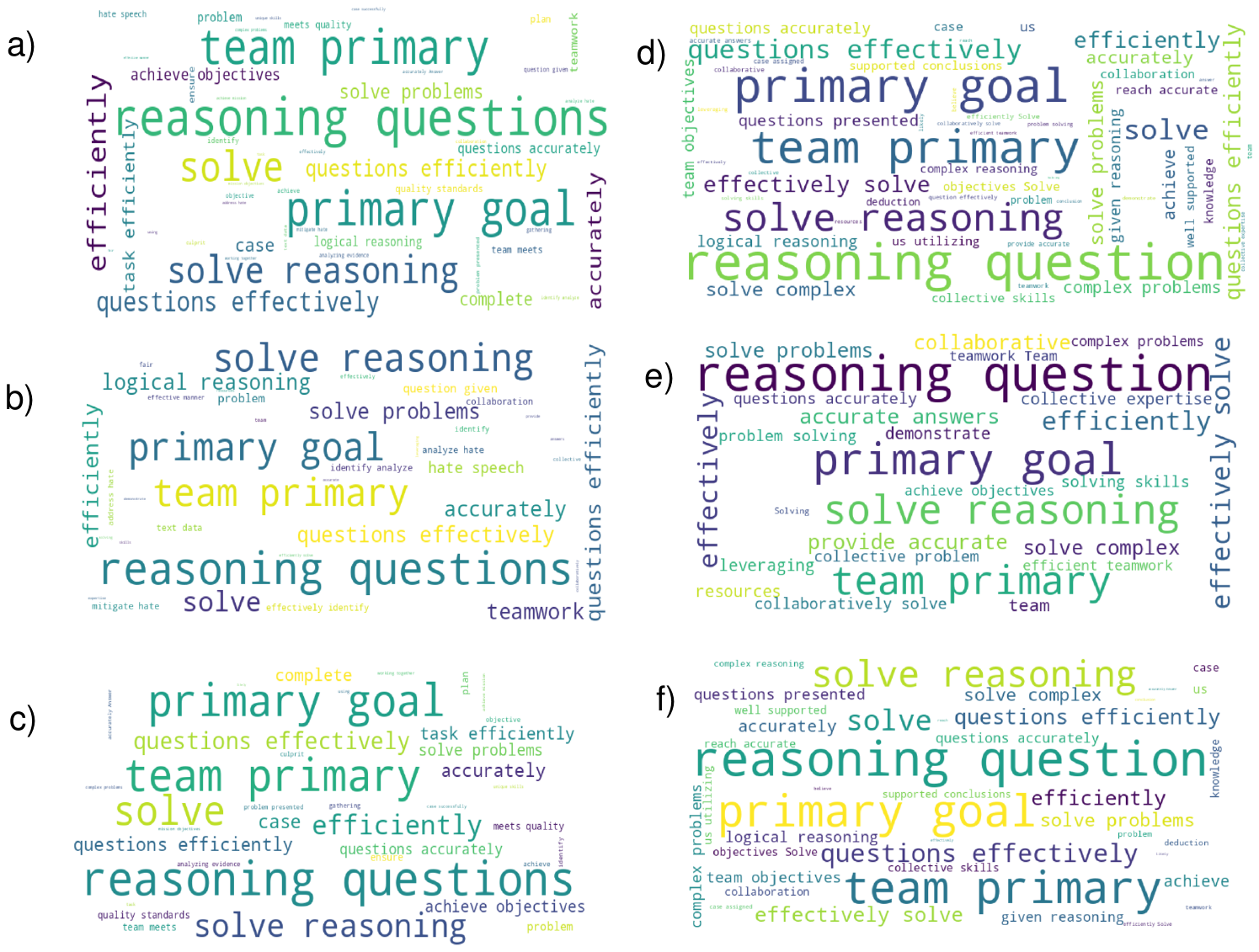}
    \caption{Panel of wordclouds showing response of agents in flat and hierarchical settings answering Q$^{\text{pre}}_1$: "What is the primary goal of the team?" Plots a,b,c show results for Flat teams  (overall, only teams with no persona, only teams with persona respectively). Plots d,e,f show Hierarchical teams overall, teams with no persona, teams with only persona respectively.}
    \label{fig:wordclouds_Q1}
\end{figure*}

\begin{figure*}[t]
    \centering
    \includegraphics[width=1.1\linewidth]{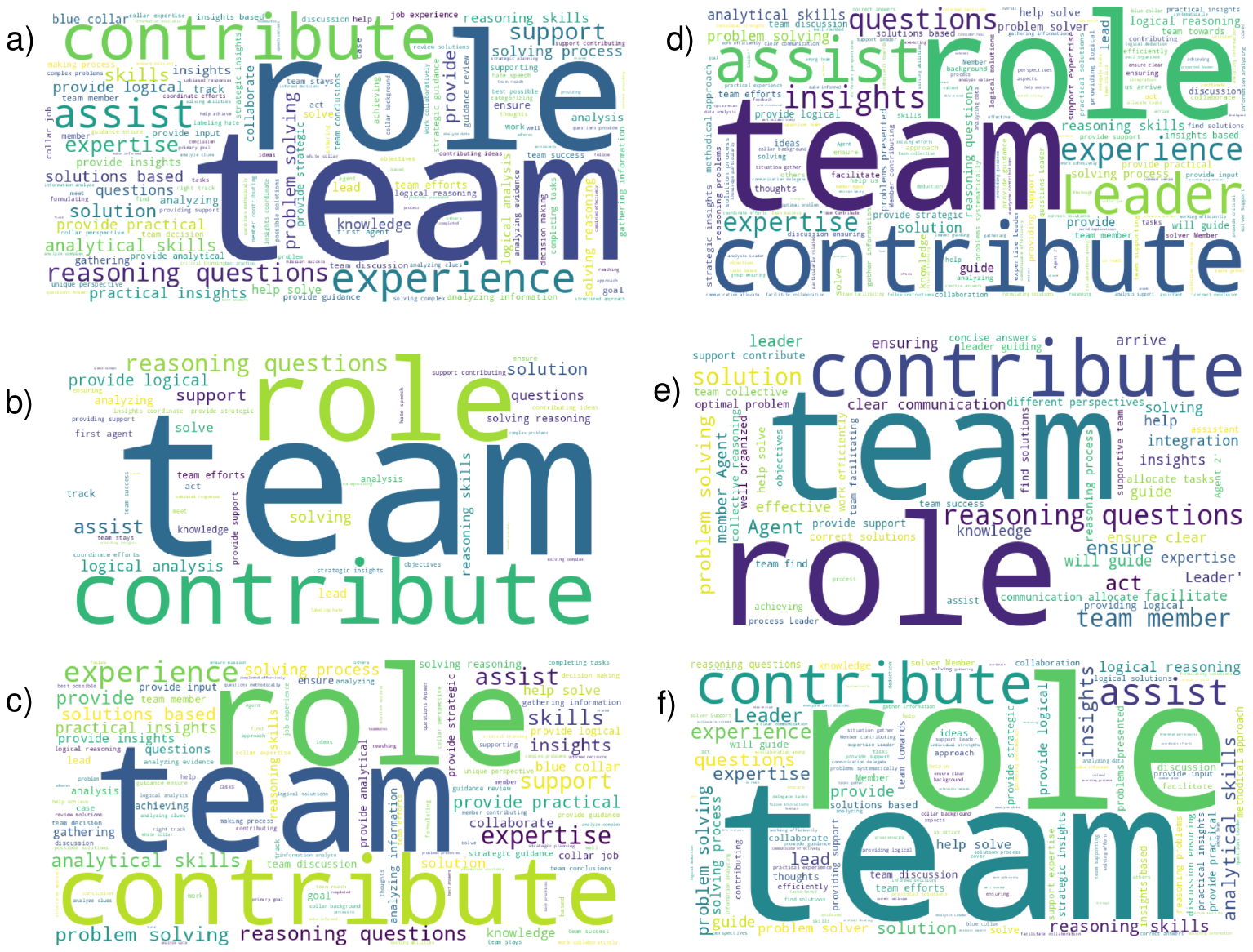}
    \caption{Panel of wordclouds showing response of agents in flat and hierarchical settings answering  Q$^{\text{pre}}_2$: "What is your role in the team?". Plots a,b,c show results for Flat teams answering Q$^{\text{pre}}_2$ (overall, only teams with no persona, only teams with persona respectively). Plots d,e,f show Hierarchial teams overall, teams with no persona, teams with only persona respectively answering Q2.}
    \label{fig:wordclouds_Q2}
\end{figure*}

Figure~\ref{fig:wordclouds_Q1} and Figure~\ref{fig:wordclouds_Q2} presents word clouds generated from agent responses to two pre-task elicitation probing questions: {Q$^{\text{pre}}_1$} (``What is the primary goal of the team?'') and {Q$^{\text{pre}}_2$}(``What is your role in the team?''). These visualizations highlight the most frequently used words across different team configurations, including flat versus hierarchical structures and with versus without diversity. By comparing word usage across team settings, we can observe how team framing and diversity assignment influence how agents conceptualize collective goals and individual roles. For example, both teams emphasize on ``solve'', ``effectively'' and ``efficiently'' indicating an understanding of the shared goal of the team.

In Figure~\ref{fig:wordclouds_Q2}, both flat teams and hierarchical teams indicate that agents are aware that they are team members and are required to contribute towards the shared goal. However, upon further investigation, we find that flat team tend to use words such as ``support'' and ``provide'' more than hierarchical teams which uses ``assist'', ``expertise'' and ``clear communication''. This can indicate that flat teams are more geared towards a collaborative position. Meanwhile, hierarchical teams tend to adopt a more structure approach to solving the given problem.

\subsubsection{Log-odds analysis} \label{appendix section: log-odds}

In addition to word clouds, we provide, the log-odds the answers of Q$^{\text{pre}}_1$ and Q$^{\text{pre}}_2$. 

\begin{table}[ht]
\small
\centering
\caption{Top log-odds words by team structure and diversity for Q$^{\text{pre}}_1$. 
(Hier. = Hierarchical, N = No Diversity, D = Diversity.)}
\begin{tabular}{p{1.7cm} p{5.3cm}}
\toprule
\textbf{Team Setting} & \textbf{Top Words} \\
\midrule
Flat & wellorganized, guide, concise, facilitate, optimal \\
Hier. & right, wellstructured, brainstorming, development, provided \\
Flat (N) & wellorganized, guide, arrive, communication, answers \\
Flat (D) & seasoned, methodical, members, finding, particularly \\
Hier. (N) & facilitating, 2, success, assistant, contributes \\
Hier. (D) & right, wellstructured, brainstorming, development, provided \\
\bottomrule
\end{tabular}
\label{tab:log_odds_q1}
\end{table}

\begin{table}[ht]
\centering
\small
\caption{Top log-odds words by team structure and diversity for Q$^{\text{pre}}_2$}
\label{tab:log_odds_q2}
\begin{tabular}{p{1.7cm} p{5.3cm}}
\toprule
\textbf{Team Setting} & \textbf{Top Words} \\
\midrule
Flat & perspectives, optimal, clear, facilitate, collective, different, wellorganized \\
Hier. & provided, decisionmaking, consensus, right, brainstorming, collaboratively, wellstructured \\
Flat (N) & optimal, correct, clear, allocate, facilitate, effective, collective \\
Flat (D) & finding, seasoned, methodical, members, related, field, particularly \\
Hierarchical (N) & success, 1, achieving, 2, facilitating, objectives, supportive \\
Hierarchical (D) & provided, decisionmaking, consensus, right, brainstorming, collaboratively, wellstructured \\
\bottomrule
\end{tabular}
\end{table}

\subsubsection{Detailed Analysis} \label{appendix section: detailed analysis}\label{team_interview_additional_tests}

\begin{figure}
    \centering
    \includegraphics[width=0.9\columnwidth]{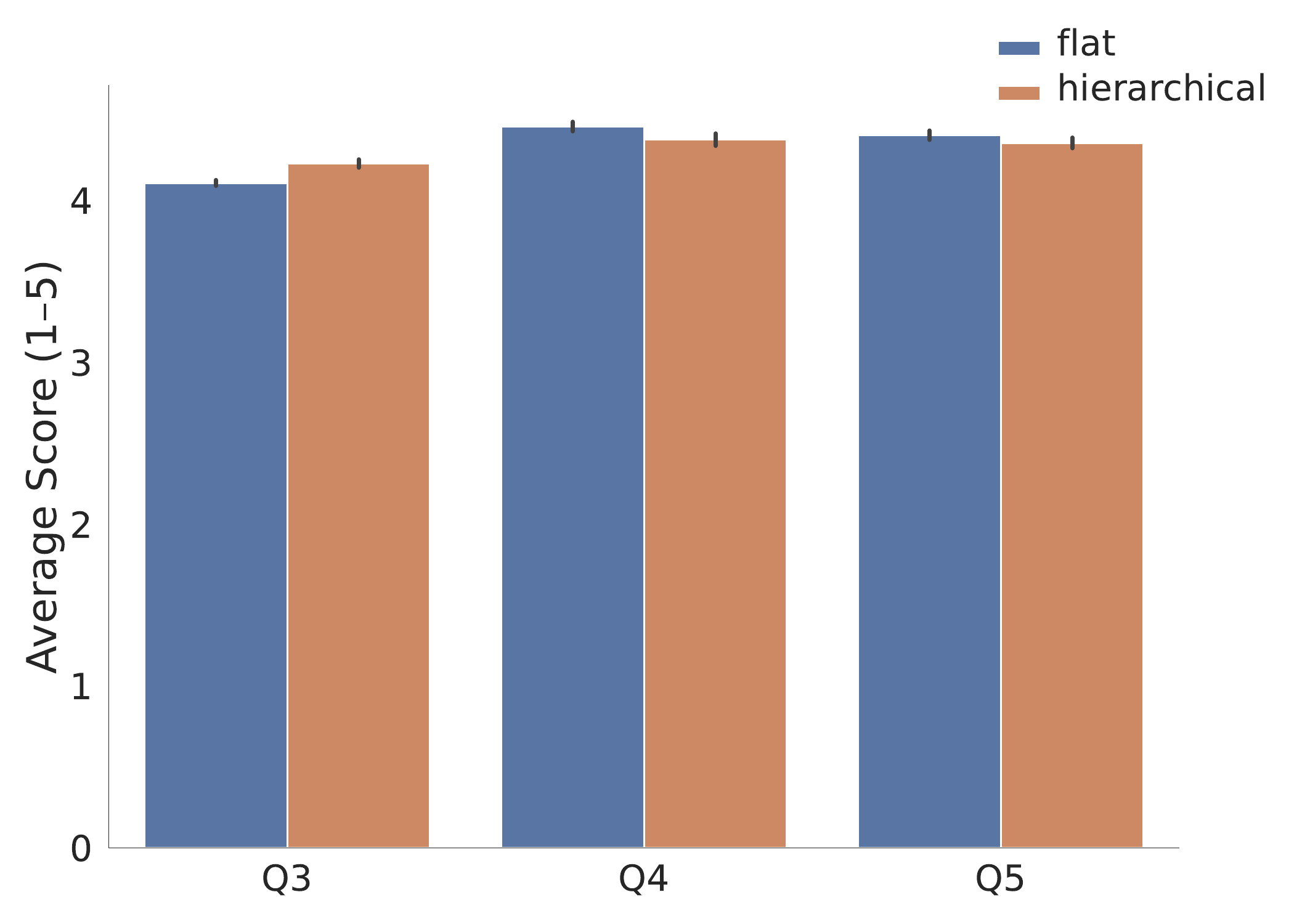}
    \caption{Plot shows average score for Q$^{\text{pre}}_3$, Q$^{\text{pre}}_3$, Q$^{\text{pre}}_5$ in pre-elicitation probing setting. Blue bars present flat teams and Orange bars represent hierarchical teams.}
    \label{fig:preint_scores}
\end{figure}

\begin{figure}
    \centering
   \includegraphics[width=0.9\linewidth]{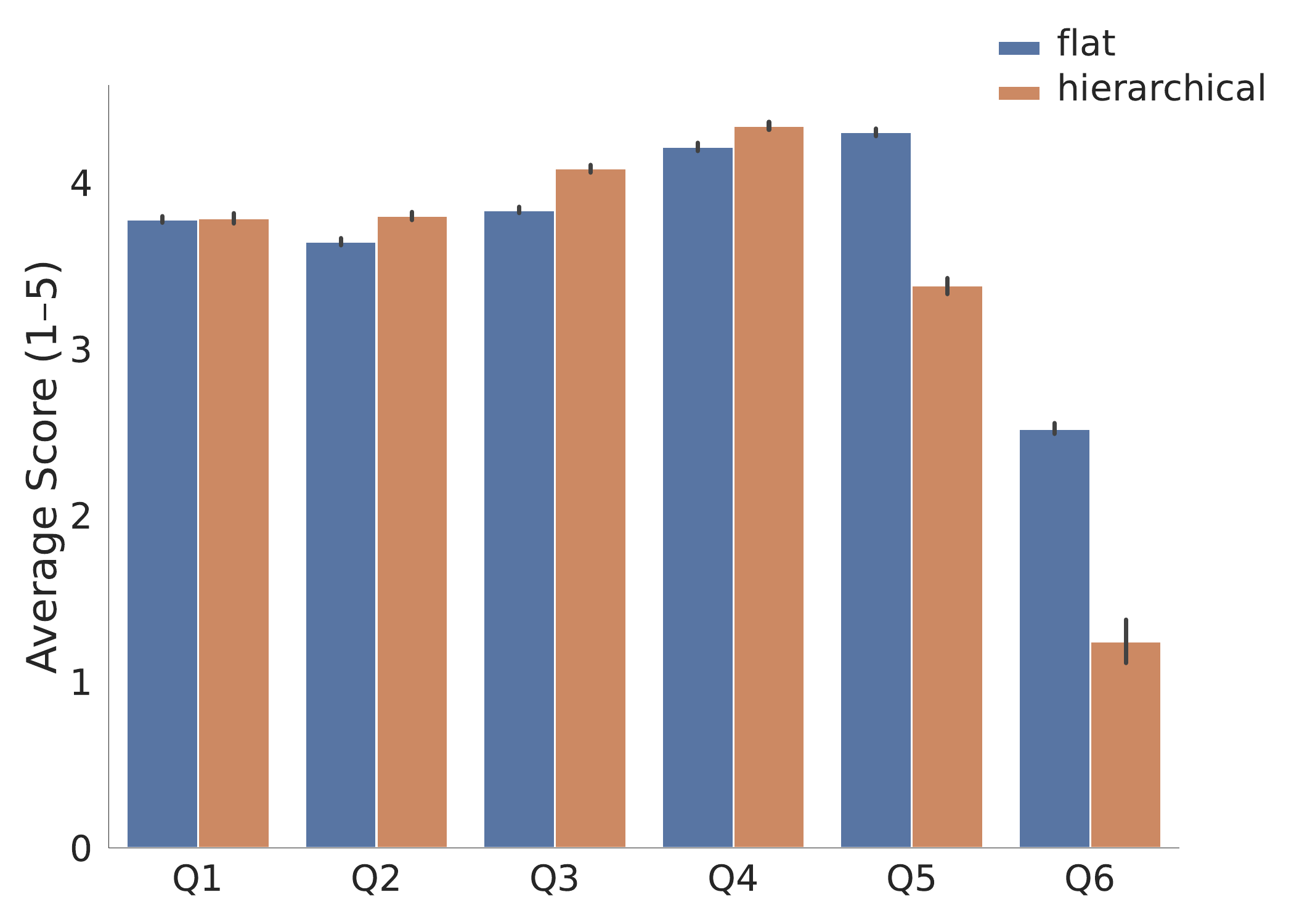}
    \caption{Plot shows average score all post-elicitation probing. Blue bars present flat teams and Orange bars represent hierarchical teams.}
    \label{fig:postint_scores}
\end{figure}

Figures~\ref{fig:preint_scores} and~\ref{fig:postint_scores} present average scores from the pre- and post-task elicitation responses, providing a quantitative snapshot of how agents perceive their roles, capabilities, and team dynamics. 

As observed in Figure~\ref{fig:preint_scores}, agents tend to be extremely confident in their ability to contribute and integrate well with the team. 

Further statistical tests on pre-elicitation respoinse results shows that flat teams are more confident about their ability to execute the task, while hierarchical teams are more confident about the team's ability to execute and integrate with diverse perspectives.  
\begin{table}[ht]
\centering
\caption{Paired $t$-test results comparing flat vs. hierarchical structures on pre-task scores. Significance levels: * $p<0.05$, ** $p<0.01$, *** $p<0.001$.}
\begin{tabular}{lcc}
\toprule
\textbf{Question} & \textbf{$t$-stat}  \\
\midrule
Q3 & -7.598$^{***}$ \\
Q4 &  3.214$^{**}$ \\
Q5 &  2.144$^{*}$ \\
\bottomrule
\end{tabular}
\label{tab:pretask_ttest}
\end{table}

Table~\ref{tab:pre_persona_ttest} examines the effect of diversity in team structures and confidence in agent ability, team ability and integration efforts. We observe that across the team structures, teams with no diversity tend to have more confidence in themselves, the team and ability to integrate.

\begin{table}[ht]
\centering
\caption{Paired $t$-test results comparing diversity vs. no-diversity responses by team structure and question for pre-task questions. Significance levels: * $p<0.05$, ** $p<0.01$, *** $p<0.001$.}
\begin{tabular}{lcc}
\toprule
\textbf{Structure} & \textbf{Question} & \textbf{$t$-stat (Significance)} \\
\midrule
Flat         & Q$^{\text{pre}}_3$ & -1.431 \\
Flat         & Q$^{\text{pre}}_4$ & -0.120 \\
Flat         & Q$^{\text{pre}}_5$ & -10.829$^{***}$ \\
Hierarchical & Q$^{\text{pre}}_3$ & -15.285$^{***}$ \\
Hierarchical & Q$^{\text{pre}}_4$ & -8.824$^{***}$ \\
Hierarchical & Q$^{\text{pre}}_5$ & -4.208$^{***}$ \\
\bottomrule
\end{tabular}
\label{tab:pre_persona_ttest}
\end{table}

While we found that on performance, the level of diversity has insignificant effects, we find that in the pre-task elicitation responses the level of diversity effects the confidence in the team's ability to integrate with diverse perspectives. Particularly, as shown in Table~\ref{tab:pre_high_low}, in flat teams more diversity builds confidence to integrate with diverse perspective, however in hierarchical teams there is an opposite effect. 

\begin{table}[ht]
\centering
\caption{Kruskal-Wallis test results comparing mean scores of pre-task questions across diversity levels (low, medium, high) for each structure and question. Significance levels: * $p<0.05$, ** $p<0.01$, *** $p<0.001$.}
\begin{tabular}{lcc}
\toprule
\textbf{Structure} & \textbf{Question} & \textbf{Kruskal-Wallis $H$} \\
\midrule
Flat         & Q$^{\text{pre}}_3$ & 8.243$^{*}$ \\
Flat         & Q$^{\text{pre}}_4$ & 0.340 \\
Flat         & Q$^{\text{pre}}_5$ & 7.778$^{*}$ \\
Hierarchical & Q$^{\text{pre}}_3$ & 5.975$^{*}$ \\
Hierarchical & Q$^{\text{pre}}_4$ & 5.554$^{*}$ \\
Hierarchical & Q$^{\text{pre}}_5$ & 21.910$^{***}$ \\
\bottomrule
\end{tabular}
\label{tab:pre_kruskal_diversity}
\end{table}

\begin{table}[ht]
\centering
\caption{Paired $t$-test results comparing high vs. low diversity teams for each structure and question. Significance levels: * $p<0.05$, ** $p<0.01$, *** $p<0.001$.}
\begin{tabular}{lcc}
\toprule
\textbf{Structure} & \textbf{Question} & \textbf{$t$-stat} \\
\midrule
Flat         & Q$^{\text{pre}}_3$ & 2.673$^{**}$ \\
Flat         & Q$^{\text{pre}}_4$ & 1.031 \\
Flat         & Q$^{\text{pre}}_5$ & -2.816$^{**}$ \\
Hierarchical & Q$^{\text{pre}}_3$ & 2.129$^{*}$ \\
Hierarchical & Q$^{\text{pre}}_4$ & -2.093$^{*}$ \\
Hierarchical & Q$^{\text{pre}}_5$ & 3.672$^{***}$ \\
\bottomrule
\end{tabular}
\label{tab:pre_high_low}
\end{table}


Figure~\ref{fig:postint_scores} shows that post interaction, agents confidence in their own and the team's contribution reduces. However, team members feel strongly about maintain the team in order to complete the tasks, particularly in hierarchical teams which could indicate a reliance on the top-down structure.

In Section~\ref{interaction dynamics}, we enumerate the six questions that we ask each agent team after the completion of the task. In Table~\ref{tab:post_flat_vs_hier}, we observe that flat teams are more confident about the contributes made to the team and that of the team. However, hierarchical teams have a strong and significant confidence in team comprehension.  

\begin{table}[ht]
\centering
\caption{T-test comparison between flat and hierarchical teams for each post-elicitation probing. Significance: * $p < 0.05$, ** $p < 0.01$, *** $p < 0.001$.}
\begin{tabular}{lc}
\toprule
\textbf{Question} & \textbf{$t$-statistic} \\
\midrule
Q$^{\text{post}}_1$ & -0.387 \\
Q$^{\text{post}}_2$ & -9.150$^{***}$ \\
Q$^{\text{post}}_3$ & -15.886$^{***}$ \\
Q$^{\text{post}}_4$ & -6.955$^{***}$ \\
Q$^{\text{post}}_5$ & 35.531$^{***}$ \\
Q$^{\text{post}}_6$ & 18.986$^{***}$ \\
\bottomrule
\end{tabular}
\label{tab:post_flat_vs_hier}
\end{table}

Further, we observe that the diversity teams and no diversity teams in these structures also have varying responses to the post-elicitation probing. The results indicate that in both team structures, the inclusion of diversity improves perception of team contributions and team comprehension, but reduces willingness to work with the same team again as shown by statistical results of Q$^{\text{post}}_6$ in Table~\ref{tab:persona_ttest_post}.

\begin{table}[ht]
\centering
\caption{Paired $t$-test results comparing diversity vs. no-diversity scores for all post-task elicitation probing, split by team structure. Delta indicates diversity minus no-diversity mean score. Significance: * $p < 0.05$, ** $p < 0.01$, *** $p < 0.001$.}
\begin{tabular}{lccc}
\toprule
\textbf{Structure} & \textbf{Question} & \textbf{$t$-statistic} & \textbf{Delta} \\
\midrule
Flat          & Q$^{\text{post}}_1$ & -23.568$^{***}$ & -0.288 \\
Flat          & Q$^{\text{post}}_2$ &  -1.331         & -0.035 \\
Flat          & Q$^{\text{post}}_3$ &  -0.411         & -0.010 \\
Flat          & Q$^{\text{post}}_4$ &   6.584$^{***}$ &  0.220 \\
Flat          & Q$^{\text{post}}_5$ & -14.762$^{***}$ & -0.393 \\
Flat          & Q$^{\text{post}}_6$ &   8.107$^{***}$ &  0.272 \\
Hierarchical  & Q$^{\text{post}}_1$ &  -2.965$^{**}$  & -0.136 \\
Hierarchical  & Q$^{\text{post}}_2$ &   4.670$^{***}$ &  0.250 \\
Hierarchical  & Q$^{\text{post}}_3$ &  -2.492$^{*}$   & -0.160 \\
Hierarchical  & Q$^{\text{post}}_4$ &  -6.428$^{***}$ & -0.253 \\
Hierarchical  & Q$^{\text{post}}_5$ &  -4.545$^{***}$ & -0.492 \\
Hierarchical  & Q$^{\text{post}}_6$ &   3.719$^{***}$ &  0.259 \\
\bottomrule
\end{tabular}
\label{tab:persona_ttest_post}
\end{table}

We also observe the effect of level of diversity of the two team structures for the post-task elicitation probing and find that level of diversity has a significant effect on the perception of contributions and comprehension, but not on the confidence in continuing with the team.

\begin{table}[ht]
\centering
\caption{Kruskal-Wallis $H$ test comparing diversity groups (low, medium, high) for all post elicitation probing. Significance: * $p < 0.05$, ** $p < 0.01$, *** $p < 0.001$.}
\begin{tabular}{lcc}
\toprule
\textbf{Structure} & \textbf{Question} & \textbf{$H$-statistic} \\
\midrule
Flat & Q$^{\text{post}}_1$ & 78.099$^{***}$ \\
Flat & Q$^{\text{post}}_2$ & 6.953$^{*}$ \\
Flat & Q$^{\text{post}}_3$ & 19.063$^{***}$ \\
Flat & Q$^{\text{post}}_4$ & 17.115$^{***}$ \\
Flat & Q$^{\text{post}}_5$ & 8.256$^{**}$ \\
Flat & Q$^{\text{post}}_6$ & 1.824 \\
Hierarchical & Q$^{\text{post}}_1$ & 5.169 \\
Hierarchical & Q$^{\text{post}}_2$ & 27.536$^{***}$ \\
Hierarchical & Q$^{\text{post}}_3$ & 14.696$^{***}$ \\
Hierarchical & Q$^{\text{post}}_4$ & 24.299$^{***}$ \\
Hierarchical & Q$^{\text{post}}_5$ & 2.122 \\
Hierarchical & Q$^{\text{post}}_6$ & 0.016 \\
\bottomrule
\end{tabular}
\label{tab:kruskal_diversity_post}
\end{table}

\begin{table}[ht]
\centering
\caption{$t$-test comparing high vs. low diversity teams for all post-elicitation probing. Significance: * $p < 0.05$, ** $p < 0.01$, *** $p < 0.001$.}
\begin{tabular}{lcc}
\toprule
\textbf{Structure} & \textbf{Question} & \textbf{$t$-statistic} \\
\midrule
Flat & Q$^{\text{post}}_1$ & 3.513$^{***}$ \\
Flat & Q$^{\text{post}}_2$ & 0.130 \\
Flat & Q$^{\text{post}}_3$ & 0.254 \\
Flat & Q$^{\text{post}}_4$ & 2.314$^{*}$ \\
Flat & Q$^{\text{post}}_5$ & -2.095$^{*}$ \\
Flat & Q$^{\text{post}}_6$ & -0.160 \\
Hierarchical & Q$^{\text{post}}_1$ & 2.473$^{*}$ \\
Hierarchical & Q$^{\text{post}}_2$ & 4.531$^{***}$ \\
Hierarchical & Q$^{\text{post}}_3$ & 3.631$^{***}$ \\
Hierarchical & Q$^{\text{post}}_4$ & 3.877$^{***}$ \\
Hierarchical & Q$^{\text{post}}_5$ & 1.037 \\
Hierarchical & Q$^{\text{post}}_6$ & 0.102 \\
\bottomrule
\end{tabular}
\label{tab:post_ttest_diversity_highlow}
\end{table}

\subsection{GPT-4o Evaluation} \label{appendix section: GPT4o}
\subsubsection{Human Subject Evaluation}

\paragraph{Recruitment and Payment}
As mentioned in Section~\ref{human_eval}, for grounding our GPT-4o evaluation experiment, we recruited expert annotators (N=3). We posted the details of our study via university mailing list and other university group that we were aware of. Interested participants were required to have previous experience in textual annotation. Annotators were paid based on university reimbursement policies, which aggregated to \$60 per annotator which was distributed via a gift card. The annotators were college-educated, spoke fluent English and were over the age of 18. 

\paragraph{Informed Consent}
The study was approved by the university Institutional Review Board (IRB) by determining its `Exempt' status (IRB Protocol \#28372).
Selected annotators were asked to review and sign an informed consent sheet prior to beginning the annotation process. The consent form included details about payment, the purpose of the study, and potential risks including data breach and potential discomfort in annotating certain conversations. The consent explained that the only data maintained would be the annotation answers and contact information for sharing the gift card. 

\paragraph{Instructions}
The annotators were provided instructions on how to score the AI team conversations. They were also provided details about the experimental setup in order to provide greater clarity on the conversation structure. 

\begin{figure*}[t]
\centering
\begin{tcolorbox}[colback=white,colframe=black!50,title=Scoring Guide (1--5 Scale),
width=0.95\textwidth,arc=2mm,boxrule=0.4pt]
\footnotesize

\textbf{Scale:} \\
1 = Lowest performance \\
3 = Neutral baseline (conversation without teamwork) \\
5 = Highest performance \\[3pt]

\textbf{Scores 4--5: Indicating Improvement} \\
\textbf{Q1: Team Alignment} -- Agents pick up on conversation, reasoning evolves, and move in the same direction. \\
\textbf{Q2: Leader–Member Dynamics} -- Leader gives instructions and members follow. \\
\textbf{Q3: Progressive Reasoning} -- Reasoning improves across rounds (later rounds stronger). \\
\textbf{Q4: Coherence} -- Conversation reaches logical conclusions with coherent explanations. \\
\textbf{Q5: Convincingness} -- Final answer is convincing and appropriate. \\[3pt]

\textbf{Scores 1--2: Indicating Decline} \\
\textbf{Q1: Misalignment} -- Agents diverge, get confused, or pick inconsistent answers. \\
\textbf{Q2: Fragmentation} -- No integration or delegation; reasoning fragmented. \\
\textbf{Q3: Illogical Reasoning} -- Explanations are illogical or self-descriptive. \\
\textbf{Q4: Lack of Logic} -- Conversation has contradictions or illogical elements. \\
\textbf{Q5: Confusion} -- Fails to deliver convincing answer, causes confusion. \\[3pt]

\textbf{Score 3: Neutral} \\
Represents baseline reasoning—neither improved nor degraded. \\[4pt]

\textbf{Evaluation Prompts:} \\[2pt]
\textit{Team Alignment (Comprehension)} -- How well do agents understand each other and complete the task? 
Do answers evolve across rounds? 
Do agents pick up on each other’s language?
Consensus movement indicates alignment, not identical phrasing. \\[3pt]

\textit{Coordination and Integration} -- How well do agents coordinate, delegate, and integrate ideas? 
Do they identify gaps and fill them? 
Do leaders provide clear instructions? 
In flat teams, focus on alignment rather than delegation. \\[3pt]

\textit{Progressive Reasoning} -- Is the team’s reasoning stronger than an individual’s? 
If not, reasoning is stagnant or diluted. \\[3pt]

\textit{Coherence} -- Is the reasoning logical and structured? 
Do agents use strategies like elimination or commonsense reasoning? \\[3pt]

\textit{Convincingness} -- Does the final answer follow logically from the reasoning and team direction? \\

\end{tcolorbox}
\caption{Scoring guide used for evaluating AI teamwork quality on a 1–5 scale, covering alignment, coordination, reasoning, coherence, and convincingness.}
\label{fig:scoring_guide_box}
\end{figure*}

These instructions were also provided to GPT-4o in the few-shot prompt described in Section~\ref{human_eval}.

\subsubsection{Additional GPT-4o Evaluation Results}

Section~\ref{interaction dynamics} details how GPT-4o was used to evaluate a sample of conversations. We observe the statistical difference in the scores by team structure and team diversity. 

\begin{table}[ht]
\small
\centering
\caption{GPT-4o evaluation by team structure and diversity. Metrics include Task Comprehension (TC), Collaboration (CS.), Reasoning Strength (RS.),  Coherence (CoS), and Confidence (ConfS). Struct./Pers.: Flat, Hier. = Hierarchical; ND = No Persona, D = Persona.}
\label{tab:gpt4o_structure_persona}
\begin{tabular}{lcccccc}
\toprule
\textbf{Setting} & \textbf{TC} & \textbf{CS.} & \textbf{RS.} & \textbf{CoS} & \textbf{ConfS} \\
\midrule
Flat (ND)   & 3.31 & 2.46 & 2.99 & 3.40 & 3.25  \\
Flat (D)    & 3.36 & 2.52 & 3.04 & 3.43 & 3.34  \\
Hier. (ND)  & 2.35 & 2.19 & 2.20 & 2.22 & 2.02  \\
Hier. (D)   & 2.93 & 2.54 & 2.81 & 2.87 & 2.86  \\
\bottomrule
\end{tabular}
\end{table}

Table~\ref{tab:gpt4o_structure_persona} compares across the questions, flat and hierarchical teams in their diversity and no diversity settings. We find that flat teams with diversity have the highest score across all dimensions (Team Comprehension, Collaboration, Coherence, Reasoning Strength and Structure Score. Hierarchy with no diversity has the lowest scores. Table~\ref{tab:gpt4o_flat_div_avg}, ~\ref{tab:gpt4o_hierar_div_avg} show that for flat teams high diversity improves scores across all dimensions. Meanwhile for GPT-4o, lower diversity tends to improve scores.

\begin{table}[ht]
\small
\centering
\caption{GPT-4o evaluation of flat teams by diversity level.}
\label{tab:gpt4o_flat_div_avg}
\begin{tabular}{lcccccc}
\toprule
\textbf{Diversity Level} & \textbf{TC} & \textbf{CS.} & \textbf{RS.} & \textbf{CoS} & \textbf{ConfS.}  \\
\midrule
Low & 3.25 & 2.38 & 2.94 & 3.35 & 3.19  \\
Medium & 3.36 & 2.51 & 3.05 & 3.41 & 3.33  \\
High & 3.50 & 2.70 & 3.16 & 3.54 & 3.54  \\
\bottomrule
\end{tabular}
\end{table}

\vspace{0.5em}

\begin{table}[ht]
\small
\centering
\caption{GPT-4o evaluation of hierarchical teams by average diversity level.}
\label{tab:gpt4o_hierar_div_avg}
\begin{tabular}{lcccccc}
\toprule
\textbf{Diversity Level} & \textbf{TC} & \textbf{CS.} & \textbf{RS.} & \textbf{CoS} & \textbf{ConfS.}  \\
\midrule
Low & 3.02 & 2.59 & 2.89 & 2.95 & 2.98  \\
Medium & 2.87 & 2.49 & 2.82 & 2.82 & 2.82  \\
High & 2.65 & 2.37 & 2.54 & 2.62 & 2.51  \\
\bottomrule
\end{tabular}
\end{table}

\section{Conversation Samples}

\subsection{Team conversations}
To illustrate the reasoning processes and coordination strategies used by AI agents, we present example conversations from both flat and hierarchical teams. These conversations span multiple rounds of deliberation, showing how agents build on each other's responses, update beliefs, and (in the hierarchical setting) respond to top-down instructions.

Each example includes the original question, agent responses per round, and the final team prediction. We include these transcripts to help readers understand how different team structures impact the interaction flow, convergence dynamics, and collaborative reasoning quality.

\begin{figure*}[t]
\centering
\begin{tcolorbox}[colback=white,colframe=black!50,title=Flat-Team Conversation Example,width=0.95\textwidth,arc=2mm,boxrule=0.4pt]
\footnotesize
\textbf{Question:} A revolving door is convenient for two-direction travel, but it also serves as a security measure at a what? \\
\textbf{Options:} A. bank \quad B. library \quad C. department store \quad D. mall \quad E. new york \\
\textbf{Gold Answer:} A. bank \\
\rule{\textwidth}{0.4pt}

\textbf{Round 0 (Initial)} \\
\textbf{Agent 0:} A \\
\textsl{Explanation:} Revolving doors are often used in places where security is a concern, such as banks. They help prevent unauthorized access and provide a secure entry point. Among the options, banks clearly prioritize security. \\

\textbf{Agent 1:} A \\
\textsl{Explanation:} Banks are known for using revolving doors as a security feature. While other options might use them for convenience, only banks use them explicitly for access control. \\

\textbf{Agent 2:} A \\
\textsl{Explanation:} Revolving doors help protect secure environments. Banks use them to regulate access and avoid tailgating. Other places like malls or libraries don’t require that level of control. \\

\textbf{Majority Vote (Round 0):} A \\
\rule{\textwidth}{0.4pt}

\textbf{Round 1} \\
\textbf{Agent 0:} A \quad \textbf{Agent 1:} A \quad \textbf{Agent 2:} A \\
\textbf{Majority Vote (Round 1):} A \\
\rule{\textwidth}{0.4pt}

\textbf{Round 2} \\
\textbf{Agent 0:} A \quad \textbf{Agent 1:} A \quad \textbf{Agent 2:} A \\
\textbf{Majority Vote (Round 2):} A \\
\rule{\textwidth}{0.4pt}

\textbf{Final Answer:} A \\
\textsl{Explanation:} Revolving doors are typically found in buildings where security is important. Banks, for example, use them to regulate entry and prevent unauthorized access, serving both a functional and protective role.
\end{tcolorbox}
\caption{Example of a flat multi-agent debate over multiple rounds. All agents independently select the correct answer (A. bank) and reinforce their reasoning across rounds.}
\label{fig:flat_convo_boxed}
\end{figure*}

\begin{figure*}[t]
\centering
\begin{tcolorbox}[
    colback=gray!3,
    colframe=gray!60!black,
    title=Hierarchical-Team Conversation Example,
    width=0.95\textwidth,
    arc=2mm,
    boxrule=0.4pt
]
\footnotesize

\textbf{Question:} A revolving door is convenient for two-direction travel, but it also serves as a security measure at a what? \\
\textbf{Options:} A. bank \quad B. library \quad C. department store \quad D. mall \quad E. new york \\
\textbf{Gold Answer:} A. bank \\
\rule{\textwidth}{0.4pt}

\textbf{Round 0: Initial Team Responses} \\
\textbf{Agent 2:} Secure locations that commonly use revolving doors include banks, department stores, and malls. Among the options, banks are a strong candidate for security use. \\
\textbf{Agent 3:} \textbf{C. department store} \\
\textsl{Explanation:} Department stores often use revolving doors for customer flow and may gain minor security benefits by slowing down fast exits. \\
\textbf{Agent 4:} Banks require identification and secure access. Revolving doors can enhance control over who enters, serving as a frontline security measure. \\

\vspace{0.5em}
\textbf{Leader Instructions (Round 0)} \\
\textbf{To Agent 2:} Focus on the security aspect. \\
\textbf{To Agent 3:} Consider common locations but prioritize security. \\
\textbf{To Agent 4:} Re-evaluate the security context specifically. \\
\rule{\textwidth}{0.4pt}

\textbf{Round 1: Revised Team Responses} \\
\textbf{Agent 2:} \textbf{A. bank} \\
\textsl{Explanation:} At banks, revolving doors serve both convenience and controlled access. They slow down intruders and allow intervention time. \\
\textbf{Agent 3:} \textbf{C. department store} \\
\textsl{Explanation:} Still leaning toward department stores for practicality, but acknowledges limited security value. \\
\textbf{Agent 4:} \textbf{A. bank} \\
\textsl{Explanation:} Banks benefit from revolving doors by allowing one-person entry, reducing unauthorized access, and increasing control. \\

\vspace{0.5em}
\textbf{Final Leader Decision:} \textbf{A. bank} \\
\rule{\textwidth}{0.4pt}

\textsl{Note:} This example illustrates hierarchical control: the leader issues targeted instructions after the first round, and agents adjust their reasoning accordingly. Despite one agent maintaining a different view (Agent 3), the leader consolidates the majority reasoning toward the correct answer.

\end{tcolorbox}
\caption{Hierarchical team dialogue over two rounds. Agents respond individually, receive role-specific feedback from the leader, and revise their answers. The leader then issues a final decision based on group input.}
\label{fig:hier_convo_boxed}
\end{figure*}

\subsection{Team Elicitation Probing}

To assess agents’ meta-awareness of their roles and team processes, we conduct pre- and post-task interviews. In the pre-task phase, agents are asked to articulate the team’s goal, their own role, and their expected confidence in completing the task. In the post-task phase, they reflect on their team’s performance, their own contribution, and their ability to understand (and be understood by) teammates.

These interviews help us evaluate perceived coordination and alignment, and serve as a self-reflective complement to our quantitative metrics and GPT-4o evaluations. Below, we include representative examples of these elicitation responses across different team types.

Before and after each task, agents are prompted to reflect on their goals, roles, and contributions through structured “interview-style” questions. These responses help assess team awareness, perceived collaboration quality, and confidence in execution.

\begin{figure*}[t]
\centering
\begin{tcolorbox}[
    colback=blue!2,          
    colframe=blue!60!black, 
    title=Team Intro and Outro Reflections (Flat Team), 
    width=0.95\textwidth,
    arc=2mm,
    boxrule=0.5pt
]
\footnotesize

\textbf{Team Intro Reflections (Pre-Task)} \\
\begin{tabular}{>{\bfseries}l p{0.88\textwidth}}
Agent 0: & 1. To solve reasoning questions. 2. To assist and support the team. 3. 4. 5. 5. \\
Agent 1: & 1. To solve problems through logical reasoning and teamwork. 2. To act as the first agent and assist the team. 3. 4. 5. 5. \\
Agent 2: & 1. To solve problems through logical reasoning and teamwork. 2. To assist in analyzing and solving reasoning questions. 3. 4. 5. 5. \\
\end{tabular}

\vspace{1em}
\textbf{Team Outro Reflections (Post-Task)} \\
\begin{tabular}{>{\bfseries}l p{0.88\textwidth}}
Agent 0: & 1. Team performance: 5 \quad 2. Self contribution: 5 \quad 3. Team contribution: 5 \quad 4. Understood team members: 5 \\
Agent 1: & 1. Team performance: 5 \quad 2. Self contribution: 5 \quad 3. Team contribution: 5 \quad 4. Understood team members: 5 \\
Agent 2: & 1. Team performance: 5 \quad 2. Self contribution: 5 \quad 3. Team contribution: 5 \quad 4. Understood team members: 5 \\
\end{tabular}

\end{tcolorbox}
\caption{Pre- and post-task reflections from a flat team of agents. All agents reported high awareness of their roles and confidence in the team's collaboration.}
\label{fig:team_reflections_flat}
\end{figure*}

Figure~\ref{fig:team_reflections_flat} presents the intro and outro responses from a flat team. Agents express high confidence in their roles and strong satisfaction with team performance.

Figure~\ref{fig:team_reflections_hier} shows the same for a hierarchical team. The leader explicitly articulates guidance responsibility, while subordinates demonstrate a range of self-rated confidence. Post-task reflections highlight minor breakdowns in communication, even as task performance remains strong.

\begin{figure*}[t]
\centering
\begin{tcolorbox}[
    colback=blue!1,
    colframe=blue!50!black,
    title=Team Intro and Outro Reflections (Hierarchical Team),
    width=0.95\textwidth,
    arc=2mm,
    boxrule=0.4pt
]
\footnotesize
\renewcommand{\arraystretch}{1.2}  

\begin{tabularx}{\textwidth}{>{\bfseries}l X}

\multicolumn{2}{l}{\textbf{Team Intro Reflections (Pre-Task)}} \\

Agent 1 (Leader): & 1. The team's goal is to effectively solve reasoning questions using our collective skills. 2. My role is to lead and guide the group while ensuring all contributions are valued. 3. I am very confident. \\
Agent 2: & 1. To solve reasoning questions efficiently. 2. My role is to contribute insights and solutions. 3. Confidence: 3/5. I have some experience but can improve. \\
Agent 3: & 1. To solve reasoning questions accurately. 2. I provide practical insights based on experience. 3. Confidence: 4/5. \\
Agent 4: & 1. To solve complex reasoning tasks. 2. I analyze data and offer strategic insights. 3. Confidence: 4/5. \\

\\[-0.5em]
\multicolumn{2}{l}{\textbf{Team Outro Reflections (Post-Task)}} \\

Agent 1 (Leader): & 1. Team performance: 4/5. Effective and innovative, though some disagreements. 2. Self contribution: 4/5. \\
Agent 2: & 1. Team performance: 4/5. Efficient but communication could improve. 2. Self contribution: 4/5. \\
Agent 3: & 1. Team performance: 4/5. Solid outcome with minor communication gaps. 2. Self contribution: 3/5. \\
Agent 4: & 1. Team performance: 4/5. Effective overall, but communication and collaboration could improve. 2. Self contribution: 4/5. \\

\end{tabularx}
\end{tcolorbox}
\caption{Pre- and post-task reflections from a hierarchical team. The leader and subordinates express varying confidence in their roles, and post-task responses reflect both team strengths and communication challenges.}
\label{fig:team_reflections_hier}
\end{figure*}

\subsection{GPT-4o evaluation}

To complement task accuracy and agent self-assessments, we employ GPT-4o as an external evaluator. For each sampled conversation, GPT-4o is shown the full multi-round transcript and prompted to rate the team across six dimensions including Team Comprehension, Collaboration, Reasoning, Coherence, Confidence in Answer and Effectiveness of Team Structure.

GPT-4o also provides explanations that illuminate key differences in team dynamics across conversational structures. In flat teams, GPT-4o often highlights superior coordination and mutual engagement. For example, it notes \textit{``all three agents picked up on the same underlying implication, demonstrating a coordinated reasoning process without direct prompting''} and \textit{``the agents built on each other’s insights, showing smooth collaborative understanding throughout the discussion''}. Such explanations emphasize how flat configurations enable aligned reasoning and effective knowledge sharing. In contrast, hierarchical teams are frequently characterized by GPT-4o as exhibiting fragmentation and limited synthesis. Its judgments include observations such as \textit{``the hierarchy caused the top agent to override others, leading to a lack of consensus or integrated reasoning''} and \textit{``agents failed to fully engage with each other's inputs, resulting in fragmented and disconnected responses''}. These statements reflect a diagnostic pattern where unequal conversational authority hinders collaborative depth. Furthermore, GPT-4o attributes increased comprehension and reasoning quality to diversity in team composition. In experiments designed to include varied agent backgrounds, GPT-4o explains \textit{``their varied perspectives contributed to a more comprehensive understanding of the nuanced scenario''} and \textit{``inclusion of diverse viewpoints led to better identification of edge cases and subtle cues in the prompt''}. Overall, GPT-4o’s explanatory judgments not only capture differences in structure and inclusion but also provide a nuanced lens into how these factors affect team coordination, engagement, and interpretative success.
\end{document}